\documentclass[9pt,5p,twocolumn]{elsarticle}

\usepackage[hidelinks,colorlinks,linkcolor=blue]{hyperref}
\usepackage{amssymb}
\usepackage{amsmath}
\usepackage{algorithm}
\usepackage{algorithmicx}
\usepackage{algpseudocode}
\usepackage{bookmark}
\usepackage{makecell}
\usepackage{orcidlink}

\journal{Image and Vision Computing}

\begin{document}

\begin{frontmatter}

\title{Dual Variational Autoencoders for Efficient Sim-to-Real Transfer in Low-Cost Robotic Navigation}

\author[label1]{\'{A}lvaro D\'{i}ez\corref{cor1}\,\orcidlink{0009-0004-6036-7733}}
\ead{alvaro.diez@ua.es}
\cortext[cor1]{Corresponding author}

\author[label1]{Fidel Aznar\,\orcidlink{0000-0003-4521-956X}}
\ead{fidel@ua.es}

\affiliation[label1]{organization={Department of Computer Science and Artificial Intelligence, University of Alicante},
            addressline={Carretera San Vicente del Raspeig}, 
            city={San Vicente del Raspeig},
            postcode={03690}, 
            state={Alicante},
            country={Spain}}

\begin{abstract}
Vision-based autonomous navigation for low-cost robots remains a fundamental challenge, primarily due to the significant gap between simulated training environments and real-world operational conditions. Direct policy transfer from simulation is often ineffective, while training exclusively on real data is impractical. We propose a hybrid transfer learning framework that effectively bridges the sim-to-real gap by combining domain randomization with feature-level domain adaptation. Our method employs a dual convolutional variational autoencoder architecture with a shared decoder, trained on an extensive set of 45225 simulated images and a minimal set of only 4556 real-world samples. This architecture learns a compact, common latent representation space that aligns the distributions of both domains. The adaptation process is further enhanced by two complementary data augmentation techniques designed to expand the limited real-world data. Experimental evaluation demonstrates that our method achieves an average success rate of almost 91\% on image classification tasks for real-world indoor navigation, significantly outperforming both simulation-only and real-world-only training. We validate these findings through a direct, real-world deployment, where the proposed policy successfully guides a low-cost robot in a reactive exploration task. Furthermore, we validate the model’s efficiency through a rigorous computational estimation, confirming its suitability for resource-constrained embedded platforms such as the Raspberry Pi 4 and NVIDIA Jetson Nano. This work presents a practical solution for developing effective and efficient navigation policies for low-cost robotic systems.
\end{abstract}


\end{frontmatter}

\section{Introduction}\label{sec:introduction}

Deep Learning (DL) [\citenum{schmidhuber_deep_2015}] has achieved remarkable success in solving a diverse array of complex tasks, particularly within vision-based applications for Artificial Intelligence (AI) and robotics. These applications span from object detection and classification [\citenum{tang_confluent_2025, deng_dsr-net_2025, liu_transifc_2025}], fine-grained recognition [\citenum{tang_learning_2022}], and human pose estimation [\citenum{liu_orientation_2023, liu_ehpe_2024, liu_ldcnet_2024}] to embodied autonomous navigation [\citenum{hoeller_learning_2021, truong_rethinking_2022}]. Across many of these fields, it is common practice to learn policies within simulated environments and subsequently transfer them to real-world platforms.

As highlighted in [\citenum{inoue_transfer_2018}], image-based learning in simulated environments offers a more straightforward training methodology compared to direct learning in reality. Physics simulators facilitate the rapid and cost-effective collection of large-scale, labeled data. In contrast, gathering extensive and properly annotated datasets in the real world can be a time-consuming, expensive, and logistically challenging endeavor.

However, the literature has also demonstrated that policies trained exclusively on simulation data often fail to generalize effectively to real-world scenarios [\citenum{shrivastava_learning_2017}, \citenum{inoue_transfer_2018}]. These difficulties primarily arise from systematic discrepancies between real and simulated environments. In the broader context of visual perception, recent studies have emphasized the critical need for robustness against real-world disturbances, such as handling occlusions [\citenum{dong_erasing_2024}], suppressing background noise and information inequality [\citenum{yan_image-specific_2024}], and defending against adversarial perturbations [\citenum{zhang_modality-specific_2025}]. Similarly, in the scientific literature, the discrepancies specifically between simulation and reality are commonly referred to by the term Reality Gap (RG) [\citenum{salvato_crossing_2021, ibarz_how_2021}].

To address these challenges, two principal sets of Transfer Learning (TL) methods have emerged: Domain Randomization (DR) [\citenum{muratore_robot_2022}] and Domain Adaptation (DA) [\citenum{wilson_survey_2020}], which can be effectively combined. Furthermore, as recently affirmed by researchers such as Azzam \textit{et al.} [\citenum{azzam_learning-based_2024}], it is also possible to effectively transfer policies from simulation to reality without employing the aforementioned explicit TL approaches, instead utilizing so-called Zero-Shot Generalization (ZSG) and Reinforcement Learning (RL) [\citenum{kirk_survey_2023}].

Considering the DR and DA transfer learning approaches, which are the focus of the present work, we investigate three core hypotheses:

\begin{enumerate}[1.]
	
	\item A DL model trained exclusively on real-world data will struggle to perform satisfactorily across novel scenarios in a real environment without an extensive, balanced training set of labeled real-world data.
	
	\item A DL model trained entirely on a large, balanced set of labeled simulation data will learn a policy with limited generalization capability to various scenarios in an equivalent real environment.
	
	\item By generating a common representation space where distributions of both extensive simulation data and limited real-world data can converge, it will be possible to mitigate the discrepancies between simulated and real environments, thereby learning a policy with improved generalization accuracy in the real world.
	
\end{enumerate}

This paper proposes a deep TL method for efficiently producing policies that can generalize and adapt to diverse situations in vision-based tasks within real environments. We focus particularly on real-time tasks requiring accurate and efficient image classification [\citenum{li_deep_2018}], such as visual navigation applications involving autonomous robots in the real world.

Specifically, our method constructs abstract low-dimensional representations from both simulation and real-world data using deep generative models. These models extract and learn meaningful features from synthetic and real images while simultaneously reducing their dimensionality. Our goal is to develop models capable of obtaining policies that can rapidly and efficiently solve various autonomous low-cost robot navigation tasks after training primarily in simulation and transferring to reality.

Our major contributions include:

\begin{itemize}
	
	\item The implementation of domain randomization by training our network with meaningful features extracted from extensive simulation data before policy transfer to reality. We collect synthetic training images using a single low-cost mobile robot in a realistic indoor environment modeled with a high-fidelity physics simulator.
	
	\item The application of domain adaptation using two convolutional Variational Autoencoders (VAEs) [\citenum{bond-taylor_deep_2022}] that share a common decoder. This decoder generates similar synthetic and pseudo-synthetic image data without requiring a direct, one-to-one correspondence between the simulated and real-world training samples, enabling successful policy generalization to novel real scenarios.
	
	\item The creation of memory-efficient image representations, making our learned policies suitable for transfer to low-cost physical robots. This is particularly relevant for autonomous reactive navigation [\citenum{patle_review_2019}, \citenum{xiao_motion_2022}] applications that use a mobile robot equipped with a single cost-effective camera.
	
	\item The validation of the developed method by transferring the proposed policy to a low-cost physical robot and executing a real-time, collision-free, map-less [\citenum{guzel_autonomous_2013}, \citenum{chang_review_2023}] visual exploration task in a real-world indoor corridor.
	
\end{itemize}

The remainder of this paper is organized as follows: Section \ref{sec:literature} reviews related work on transfer learning and its main approaches, as well as specific studies on autonomous robot navigation applications in the real world. Section \ref{sec:methodology} details our proposed transfer learning method, including data acquisition, training methodology, and data augmentation techniques. Section \ref{sec:experiments} presents and analyzes our experimental results, including a proof-of-concept validation performed in the real world using a mobile robot with limited computational resources. Finally, Section \ref{sec:conclusions} addresses our conclusions and directions for future research.

\section{Related Work}\label{sec:literature}

To situate our proposed hybrid TL framework, this section provides a concise overview of the prevailing strategies for bridging the simulation-to-reality gap in robotics. We first review foundational research in the two principal families of transfer learning that our method combines: domain randomization and domain adaptation. Subsequently, we examine an alternative paradigm, zero-shot generalization, to provide a comprehensive theoretical context. Finally, we survey pertinent applications in real-world robotic navigation that share objectives with our study, thereby contextualizing our approach for achieving efficient sim-to-real transfer on resource-constrained platforms.

\subsection{Domain Randomization}\label{subsec:randomization}

Domain randomization encompasses methods used to train more generalizable models in a source domain (\textit{e.g.}, a simulated environment) for effective prediction across systematically different target domains (\textit{e.g.}, distinct real-world environments) [\citenum{sadeghi_cad2rl_2017}]. During training, various configurations are randomized in simulation, including object placement, illumination conditions, and surface textures.

Domain randomization has emerged as a pivotal strategy for mitigating the reality gap, functioning as a regularization technique that prevents a DL model from overfitting to individual simulation instances [\citenum{muratore_robot_2022}]. The underlying hypothesis is that if the variability in simulation is sufficiently extensive, the learned policy will become robust enough to be transferred directly to a physical robot without requiring any real-world data for fine-tuning [\citenum{tobin_domain_2017}].

Two seminal works were instrumental in establishing this TL approach in robotics. In 2017, Sadeghi and Levine [\citenum{sadeghi_cad2rl_2017}] leveraged domain randomization for real-world navigation tasks. Their method generated a collision avoidance policy using a Convolutional Neural Network (CNN) [\citenum{lecun_deep_2015}] trained through deep reinforcement learning. Their model processed raw Red, Green, Blue (RGB) camera images to predict collision probability, enabling velocity command determination for collision-free indoor flight of a physical quadrotor.

Unlike our TL approach, which incorporates real-world data, their navigation policy relied entirely on synthetic training image data from a simulated environment. Their method did not require particularly high-fidelity simulation or extreme realism. Instead, Sadeghi and Levine extensively randomized rendering settings for their synthetic dataset, forcing the network to handle diverse obstacle appearances and lighting conditions. This approach made their learned high-dimensional representations invariant to surface appearance, enabling the model to learn geometric features for robust open space detection across different real indoor environments. The initial training phase of our proposed method is fundamentally rooted in this principle of procedural randomization.

Concurrently, Tobin \textit{et al.} [\citenum{tobin_domain_2017}] popularized domain randomization for robotic manipulation tasks. One year later, in [\citenum{tobin_domain_2018}], they proposed a novel deep generative architecture for learning grasp sampling distributions across diverse objects. Their network was trained exclusively on unrealistic procedurally generated objects from physics simulation. The architecture developed by Tobin \textit{et al.} employed an autoregressive model [\citenum{bond-taylor_deep_2022}] that processed embedding representations mapped from target object observations through an image representation module. The model generated distributions over possible grasps using additional depth images captured by a simulated robot's hand camera. The architecture aimed to select optimal grasps from the sampled possibilities. Their network achieved 92\% success rate on unseen realistic objects during simulation testing and 80\% success rate on real-world grasp attempts.

Like our DR method, the previous works are based on the so-called static domain randomization, which typically relies on uniform sampling from a ﬁxed domain parameter distribution. This particular approach usually implements sim-to-real transfer using any real-world data. Although it represents a technique that can effectively, quickly, and relatively easily bridge the reality gap, learned policies tend to perform worse in the target domain than those produced adaptively. [\citenum{muratore_robot_2022}].

Adaptive domain randomization employs different strategies that refine the simulation distribution during the learning process, introducing efficiency and higher fidelity by closing the sim-to-real loop. On the one hand, curriculum-based approaches, such as automatic domain randomization, seeks to increase the sample efficiency by introducing a schedule of increasing difficulty [\citenum{muratore_robot_2022}]. On the other hand, active domain randomization aims to improve training efficiency by actively searching for the most informative or challenging simulation parameters for the current policy, thereby focusing computational effort on those hard cases that are most likely to improve robustness [\citenum{mehta_active_2019}].

The most advanced methods belonging to adaptive domain randomization close the sim-to-real loop by using data from the real world to directly guide the simulation process. This represents a fundamental conceptual shift from generalization to adaptation. A notable example is the task-driven framework called AdaptSim, proposed by Ren \textit{et al.} [\citenum{ren_adaptsim_2023}]. They meta-learned an adaptation policy through reinforcement learning using a simulator to adjust the simulation parameter distribution based on its performance on the target task. Subsequently, the policy was deployed in the real world, utilizing a few interactions to find the optimal simulation configuration for re-training.

Our proposed approach offers a compelling and practical alternative to online, simulator-centric adaptation schemes. Instead of adapting the simulator to match reality, our method adapts the model's perception of reality by fine-tuning a pre-trained variational autoencoder using a small batch of real data. This model-centric adaptation strategy is offline and highly data-efficient, making it particularly suitable for low-cost platforms where continuous real-world interaction for iterative simulator tuning is impractical or unsafe.

\subsection{Domain Adaptation}\label{subsec:adaptation}

Whereas domain randomization operates exclusively on the source domain (\textit{e.g.}, a simulated environment) to foster generalization, domain adaptation is a transfer learning paradigm that actively leverages data from the target domain (\textit{e.g.}, a real-world environment) to align the distributions between the source and target environments [\citenum{wilson_survey_2020}].

Contemporary DA techniques for sim-to-real transfer can be broadly categorized into two main families: pixel-level adaptation, which modifies the visual appearance of source images, and feature-level adaptation, which aligns the underlying representations learned by the model [\citenum{wilson_survey_2020}]. Our proposed method employs generative feature-level domain adaptation to fine-tune a robust model initially pre-trained with domain randomization.

Pixel-level adaptation methods aim to bridge the reality gap by translating synthetic source images to appear as if they were drawn from the target domain's distribution. This is predominantly achieved using Generative Adversarial Networks (GANs) [\citenum{bond-taylor_deep_2022}], particularly through unpaired image-to-image translation architectures like CycleGAN [\citenum{zhu_unpaired_2017}].

However, a critical challenge with this approach is the potential for semantic inconsistency. A GAN optimized solely for visual realism may produce photo-realistic images at the expense of altering or even removing task-critical content, such as obstacles or key objects, leading to catastrophic policy failure [\citenum{rao_rl_2020}]. To mitigate this, task-aware extensions have been developed. A notable example is RL-CycleGAN, proposed by Rao \textit{et al.} [\citenum{rao_rl_2020}], which introduces an additional reinforcement learning-based consistency loss. Our proposed method circumvents the training instability and semantic consistency issues inherent in GANs by operating at the feature level, which offers a more stable and direct approach to aligning domain-invariant structural information.

Instead of altering input images, feature-level adaptation methods seek to learn a mapping that projects both source and target data into a common, domain-invariant representation space where their distributions are aligned. Within this shared feature space, a classifier trained on labeled source data can generalize effectively to unlabeled target data [\citenum{wilson_survey_2020}]. These methods are primarily divided into two families: statistical and adversarial.

Statistical methods directly minimize a distance metric between the feature distributions of the two domains. A prominent example is Deep Correlation Alignment (Deep CORAL), which introduces a CORAL loss to align the second-order statistics (\textit{i.e.}, covariances) of the source and target feature distributions within one or more layers of a deep network [\citenum{sun_deep_2016}].

In contrast, adversarial methods employ a domain discriminator in a zero-sum game against the feature extractor. The discriminator is trained to distinguish the origin of the features (\textit{i.e.}, the source domain or the target domain), while the feature extractor is trained to produce features that fool the discriminator, thereby forcing the distributions to become indistinguishable [\citenum{wilson_survey_2020}]. A seminal work in this area is the Domain-Adversarial Neural Network (DANN), which uses a novel Gradient Reversal Layer (GRL) to enable end-to-end adversarial training [\citenum{ganin_domain-adversarial_2016}]. These foundational methods, many of which are implemented in standard libraries like Awesome Domain Adaptation Python Toolbox\footnote{https://adapt-python.github.io/adapt/} (ADAPT), form the basis for modern feature-level adaptation.

Our work is situated within the domain of generative feature-level adaptation, which leverages deep generative models to learn a common latent space. The choice between a VAE and a GAN for this purpose involves a fundamental trade-off. While GANs can produce sharp, high-fidelity images, their adversarial training is notoriously unstable. VAEs, conversely, offer highly stable training and produce a continuous, structured latent space well-suited for control tasks, though their reconstructions tend to be blurry [\citenum{bond-taylor_deep_2022}].

\subsection{Zero-Shot Generalization}\label{subsec:generalization}

An alternative paradigm to explicit transfer learning is zero-shot generalization, which aims to train a policy entirely in simulation that can be deployed directly in the real world without any fine-tuning or adaptation on target domain data [\citenum{kirk_survey_2023}]. This zero-shot sim-to-real transfer is often pursued through hybrid approaches that combine reinforcement learning with domain randomization to create a robust policy [\citenum{xiao_flying_2023}]. However, a distinct line of research, and the focus of this subsection, investigates the feasibility of achieving ZSG by relying exclusively on RL, foregoing explicit DR or domain adaptation.

One prominent strategy for RL-only ZSG is to invest in creating a high-fidelity digital twin of the physical system. The underlying hypothesis is that if the simulator's dynamics are sufficiently accurate, the reality gap becomes negligible, allowing a standard RL policy to transfer directly. A compelling demonstration of this approach is presented by Azzam \textit{et al.} [\citenum{azzam_learning-based_2024}], who developed a navigation and collision avoidance policy for an Unmanned Aerial Vehicle (UAV). Their framework employs a hierarchical control structure where the RL agent learns high-level velocity commands, which are then executed by a low-level controller.

The key to their success in achieving direct sim-to-real transfer without any randomization was the meticulous system identification of the UAV's dynamic model using a deep neural network-based method. By ensuring the simulated dynamics closely mirrored reality, the policy learned by Azzam \textit{et al.} achieved a 90\% success rate in real-world experiments. This result is remarkably comparable to the 90.88\% classification accuracy achieved by our proposed method on our real-world test dataset.

Counter-intuitively, an alternative approach posits that lower, rather than higher, simulation fidelity can lead to superior sim-to-real transfer. This hypothesis, explored by Truong \textit{et al.} [\citenum{truong_rethinking_2022}], argues that high-fidelity physics simulations can be computationally slow and may cause the policy to overfit to simulator-specific inaccuracies that do not exist in the real world. Their solution is to strategically abstract away the complex dynamics. Truong \textit{et al.} propose a hierarchical decomposition where a high-level navigation policy is trained with RL in a fast, low-fidelity kinematic simulation. This policy learns to output desired velocity commands, which are then passed to the robot's native, potentially black-box, low-level controller for execution in the real world.

The approaches reviewed above highlight a fundamental trade-off in sim-to-real transfer. End-to-end RL systems for ZSG are predicated on minimizing the reality gap \textit{a priori}, either by investing in robot-specific system identification [\citenum{azzam_learning-based_2024}] or by abstracting away complex dynamics [\citenum{truong_rethinking_2022}]. Both strategies can be brittle. The former is sensitive to any unmodeled dynamics or changes in the physical robot over time, while the latter is limited by the expressive capacity of the abstracted action space [\citenum{truong_rethinking_2022}].

In contrast, our proposed method acknowledges the inherent imperfections of any simulator. It adopts a hybrid, few-shot sim-to-real strategy that is more robust and flexible. By first leveraging domain randomization, it learns a generalized feature representation that is not overfitted to a single simulation instance. Subsequently, it performs a highly efficient, model-centric adaptation by fine-tuning a pre-trained variational autoencoder using a small batch of unlabeled real images.

A third and rapidly emerging pathway to generalization in robotics is through the use of large-scale, pre-trained foundation models. Models such as RT-2 [\citenum{zitkovich_rt-2_2023}] derive their remarkable generalization capabilities not from mastering simulation physics, but from being pre-trained on vast, internet-scale datasets of real-world images and text. This allows them to acquire a broad semantic understanding of objects, concepts, and actions, enabling impressive zero-shot and few-shot performance on a wide range of manipulation and navigation tasks. However, this approach introduces a critical trade-off. While foundation models excel at open-vocabulary reasoning, they are computationally massive, with memory and processing requirements that are prohibitive for the low-cost, resource-constrained robotic platforms [\citenum{zitkovich_rt-2_2023}] that are the focus of our research.

\subsection{Real-World Autonomous Navigation Systems}\label{subsec:navigation}

Recent research in autonomous navigation has yielded a diverse array of specialized solutions, each tailored to specific application domains and operational constraints. In the automotive sector, studies range from the development of robust fuzzy logic controllers for enhancing electric vehicle stability and power quality [\citenum{sahoo_robust_2020}] to the design of neural networks for the semantic segmentation of radar point clouds [\citenum{cennamo_neural_2021}].

Biologically-inspired methods have also gained prominence in robotics research. Aljalbout \textit{et al.} [\citenum{aljalbout_task-independent_2020}], for instance, developed a generalizable framework using spiking neural networks to instantiate central pattern generators for legged locomotion across different robot morphologies with minimal design effort. More aligned with our focus on mobile robots, Chen \textit{et al.} [\citenum{chen_path_2022}] proposed the Cuckoo-Beetle Swarm Search (CBSS) algorithm for path planning [\citenum{patle_review_2019}] in heterogeneous robot teams. Their bio-inspired optimization method developed a navigation system for obstacle avoidance in indoor environments.

Consistent with the methodology of this study, Chen \textit{et al.} first validated their system in simulation before deploying it on real Unmanned Ground Vehicles (UGVs). However, a key distinction lies in the sensor modality; their method relied on a Light Detection and Ranging (LiDAR) sensor [\citenum{kolar_survey_2020}] for environmental perception. While LiDAR provides high-precision depth information, its associated cost and weight can be prohibitive for the low-cost platforms central to our research.

More closely aligned with our work's emphasis on minimalist sensing, Othman and Rad [\citenum{othman_doorway_2020}] presented a visual system belonging to an end-to-end navigation strategy for behavior-based robots [\citenum{yeon_implementation_2015}], focusing on doorway detection using only monocular RGB camera input. Their approach employed a convolutional neural network to classify scenes as containing an open door or not, guiding navigation through depth estimation and obstacle detection modules. The system was validated on a NAO humanoid robot, using a single camera to demonstrate the potential of minimal sensor configurations for executing complex navigation subtasks.

\section{Sim-to-Real Transfer with Dual Variational Autoencoders}\label{sec:methodology}

This section details our proposed hybrid transfer learning method, which is designed to bridge the simulation-to-reality gap for vision-based robotic navigation. Our framework integrates two principal strategies: domain randomization and domain adaptation. The core of our approach is a dual convolutional variational autoencoder architecture that shares a common decoder. This structure learns a shared latent representation space where the feature distributions of simulated and real-world images can converge.

To further enhance this alignment, we employ two distinct data augmentation techniques to synthesize additional training samples from the limited real-world data, thereby improving the model's generalization capabilities in the target domain. The following subsections describe our data acquisition protocol, the architecture for generating the shared representation space, and the data augmentation strategies used to refine it.

\subsection{Data Acquisition for Hybrid Sim-to-Real Transfer}\label{subsec:acquisition}

The successful implementation of our hybrid TL method requires two distinct datasets: an extensive, labeled synthetic dataset from the source domain (\textit{i.e.}, simulation) and a small, labeled real-world dataset from the target domain (\textit{i.e.}, reality).

\begin{figure*}[h!]
	\centering
	\includegraphics[width=\textwidth]{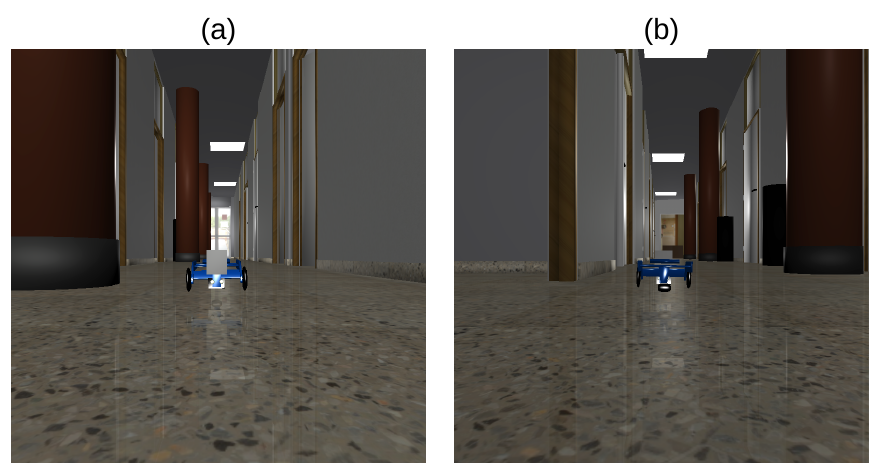}
	\caption{The simulated indoor environment used for synthetic data acquisition. This is a high-fidelity representation of a specific corridor located in the Department of Computer Science and Artificial Intelligence at the University of Alicante. (a) View of the low-cost robot from the corridor's entrance. (b) View of the low-cost from the corridor's end.}\label{fig:environment}
\end{figure*}

To generate the synthetic source dataset, we employed a custom high-fidelity physics simulator to model an indoor corridor environment, shown in Fig. \ref{fig:environment}(a) and Fig. \ref{fig:environment}(b), and a Makeblock mBot Ranger\footnote{https://www.makeblock.com/pages/mbot-ranger-robot-building-kit} robot equipped with a monocular RGB camera [\citenum{kolar_survey_2020}].

This phase implements a static domain randomization strategy. During data collection, we expose the virtual robot to diverse scenarios featuring various fixed obstacles (\textit{e.g.}, walls, columns, and door frames), varying illumination conditions, and elements with realistic textures, such as the floor surface. Our robot encounters these varying conditions while performing collision-free exploration tasks using a depth sensor [\citenum{kolar_survey_2020}].

This process yielded a large-scale dataset of 45225 labeled synthetic images, designed to train a policy that is inherently robust to a wide range of visual appearances. Our simulator, built upon the Planck.js\footnote{https://piqnt.com/planck.js/} 2D physics engine, is designed to facilitate the rapid and scalable collection of labeled data for various robotic learning tasks.

For the target domain, we used the identical physical Makeblock mBot Ranger robot to capture a significantly smaller dataset of 4556 images within the corresponding real-world indoor corridor. This limited dataset intentionally reflects the practical constraints of time, cost, and labor associated with real-world data gathering. These real images are crucial for the domain adaptation phase, where they are used to fine-tune the model pre-trained on simulated data, as detailed in the subsequent subsections.

\subsection{Dual Encoder-Decoder Architecture for Shared Latent Space Learning}\label{subsec:training}

\begin{figure*}[h!]
	\centering
	\includegraphics[width=\textwidth]{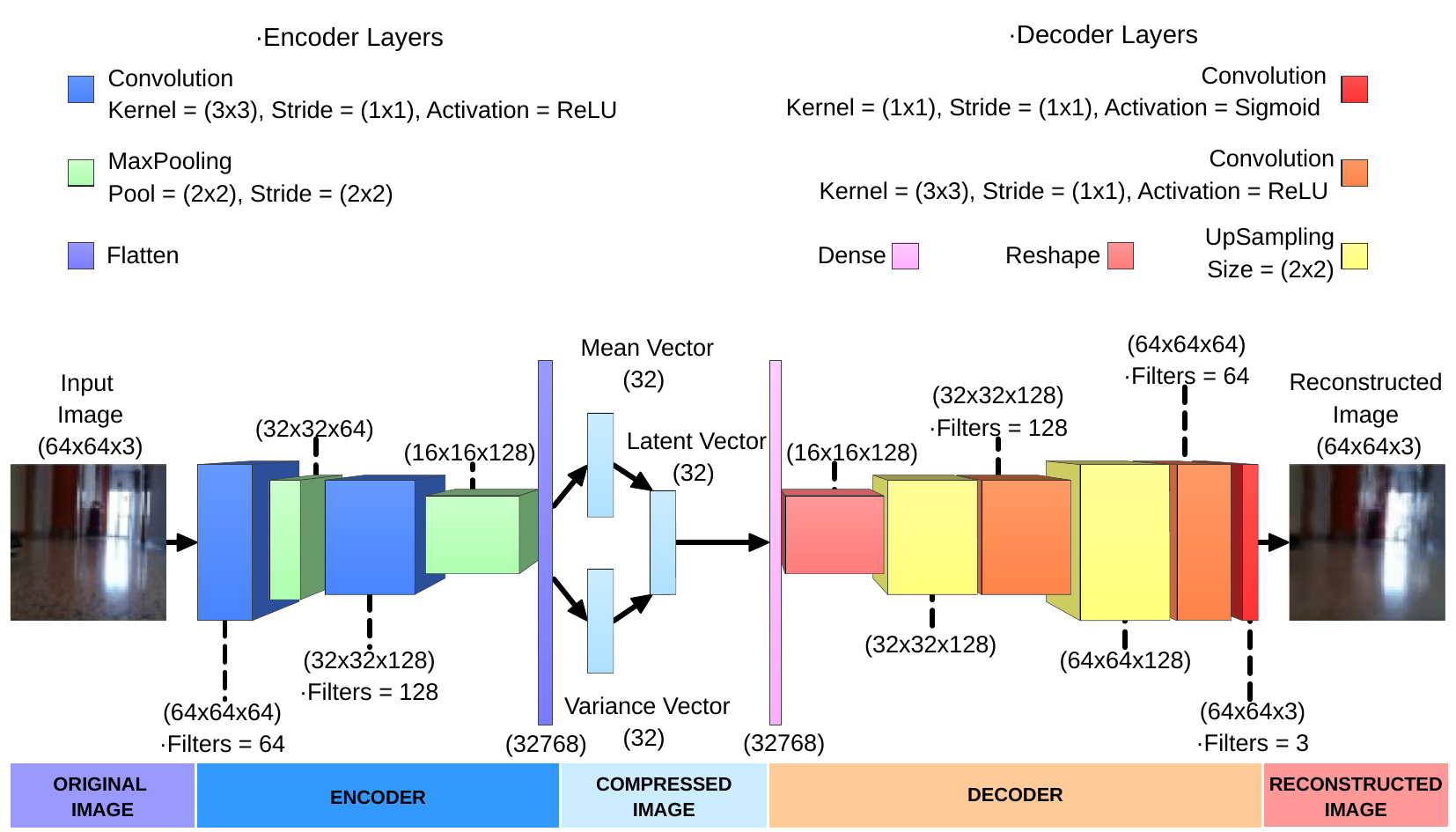}
	\caption{Layer structure of our variational autoencoders. Each model comprises a convolutional encoder that compresses a 64x64 RGB input image into a 32-dimensional latent vector via variational inference, and a convolutional decoder that reconstructs the image from this latent representation.}\label{fig:architecture}
\end{figure*}

Our method for generating the common representation space is centered on a deep neural network architecture composed of two convolutional variational autoencoders, denoted as $G_{S}$ and $G_{R}$, which share a common decoder. The layer configuration for these VAEs is detailed in Fig. \ref{fig:architecture}. The architecture is intentionally kept compact to ensure that the learned policies are memory-efficient and suitable for real-time deployment on low-cost robotic platforms. As demonstrated in the literature [\citenum{kuroswiski_autonomous_2018}, \citenum{aldegheri_fast_2018}, \citenum{le_low-cost_2020}], memory efficiency and processing speed are crucial for low-cost robotics due to inherent memory, processing, and power constraints.

Our VAEs learn to extract meaningful features from both synthetic image data $S$ (\textit{i.e.}, our source domain) and real image data $R$ (\textit{i.e.}, our target domain). These deep neural networks generate abstract low-dimensional representations defined as $Z = \{z_i\}_{i = 1}^{N}$. Each representation is a latent vector, $z_i = \{z_k\}_{k = 1}^{K}$ (\textit{e.g.}, a 32-element representation where $K=32$), corresponding to a particular image instance from an input set, $X = \{x_i\}_{i = 1}^{N}$.

An encoder network, defined by an approximate posterior distribution $q_\phi(z_i \vert x_i)$, maps an input image $x_i$ to a low-dimensional latent vector $z_i$. As explained in [\citenum{bank_autoencoders_2023}], this is achieved by sampling from a learned Gaussian distribution:
\begin{equation}
	z_i = h(x_i)\varepsilon + f(x_i), \quad \varepsilon \sim \mathcal{N}(0, I),\label{eqn:sampling}
\end{equation}
where $f(x_i)$ and $h(x_i)$ represent the mean and covariance of the posterior, and $\epsilon$ is sampled from a standard normal distribution $\mathcal{N}(0,I)$. This equation helps determine our loss function for training the convolutional variational autoencoders. Drawing inspiration from [\citenum{bank_autoencoders_2023}], we define our loss function as:
\begin{equation}
	\mathbb{L} = \sqrt{\frac{1}{N}\sum\limits_{i = 1}^{N}(x_i - \tilde{g}(z_i))^{2}} + \mathcal{D}_{KL}(\mathcal{N}(f(x_i), h(x_i)), \mathcal{N}(0, I)).\label{eqn:loss}
\end{equation}

The first term is the root mean squared error between the input images $\{x_i\}_{i = 1}^{N}$ and their reconstructed counterparts, $\tilde{x}_i = \tilde{g}(z_i)$. The latter represents the inverse function of $z_i = {g}(x_i)$, namely the compression of input images $X$ into their respective latent vectors $Z$. To quantify the reconstruction loss, it is common to use directly the mean squared error instead of the root mean squared error. The latter allows us to obtain lower values for our loss function, contributing to greater training stability.

The second term is the Kullback-Leibler (KL) divergence [\citenum{bank_autoencoders_2023}], which regularizes the encoder by encouraging the learned latent distribution to approximate a standard normal prior, thus preventing overfitting and promoting generalization.

Our training method ultimately aims to compute a conditional distribution $p(x_i^{PS} \vert x_i^{R})$, which can generate pseudo-synthetic images $\{x_i^{PS}\}_{i = 1}^{N}$ from real-world inputs $\{x_i^{R}\}_{i = 1}^{N}$. This distribution creates our common representation space, enabling us to reduce the domain discrepancy between simulation and reality.

Drawing from [\citenum{inoue_transfer_2018}], we combine a specific convolutional encoder and decoder to obtain the aforementioned ultimate distribution. The encoder network, represented as an approximate distribution $q_\varphi^{R}(z_i \vert x_i^{R})$, learns from both simulation and real data. The decoder network, defined as a conditional distribution $p_\theta(x_i^{S} \vert z_i)$, functions as a generative model trained exclusively on simulation images.

Our training process implements the hybrid DR-DA strategy in two stages. First, we train the initial convolutional VAE, $G_{S}$, on our extensive synthetic dataset, $X_{S} = \{x_i^{S}\}_{i = 1}^{N}$, created through the process detailed in Subsection \ref{subsec:acquisition}. This initial training minimizes the following lower bound:
\begin{equation}
	\begin{split}
		\mathcal{L}_{G_S}(\theta, \phi; x_i^{S}) = & -\mathcal{D}_{KL}(q_\phi^{S}(z_i \vert x_i^{S})\|p_\theta(z_i))\\ & + \mathbb{E}_{q_\phi^{S}(z_i \vert x_i^{S})}[\log p_\theta(x_i^{S} \vert z_i)].\label{eqn:simulation}
	\end{split}
\end{equation}

This domain randomization stage allows the shared decoder, $p_\theta(x_i^{S} \vert z_i)$, to learn a rich generative model of the world from diverse simulated scenarios. Second, we perform feature-level domain adaptation. We initialize the second convolutional VAE, $G_{R}$, with the weight parameters $\phi$ and $\theta$, previously learned for the encoder and decoder of $G_{S}$, respectively.

We then freeze the parameters $\theta$ of the shared decoder and fine-tune only the encoder's parameters $\varphi$ using our limited real-world dataset, $X_{R} = \{x_i^{R}\}_{i = 1}^{M}$. This fine-tuning process minimizes the following lower bound:
\begin{equation}
	\begin{split}
		\mathcal{L}_{G_R}(\varphi; x_i^{R}) = & -\mathcal{D}_{KL}(q_\varphi^{R}(z_i \vert x_i^{R})\|p_\theta(z_i))\\ & + \mathbb{E}_{q_\varphi^{R}(z_i \vert x_i^{R})}[\log p_\theta(x_i^{PS} \vert z_i)].\label{eqn:reality}
	\end{split}
\end{equation}

This last step adapts the encoder, yielding $q_\varphi^{R}(z_i \vert x_i^{R})$, to map real images into the shared latent space learned from simulation, thereby bridging the visual domain gap.

\subsection{Enhanced Domain Alignment via Data Augmentation}\label{subsec:augmentation}

A primary challenge in our approach is the limited size of the real-world dataset, which can constrain the effectiveness of the domain adaptation phase. To further bridge the sim-to-real gap and enhance the robustness of our model, we introduce the two complementary data augmentation [\citenum{shorten_survey_2019}] techniques promised earlier to expand the real-world training data and refine the unsupervised fine-tuning of the pseudo-synthetic autoencoder $G_{R}$. This process is outlined in Algorithm \ref{alg:training}.

Our first technique involves applying a set of standard image-space transformations, $O = \{o_{crop}, o_{brightness}, o_{contrast}, o_{saturation}\}$, to the original real-world dataset $X_{R}$. These transformations include random cropping, as well as altering image brightness, contrast, and saturation by randomly perturbing individual pixel values. This process yields a new image dataset $X_{D}$:
\begin{equation}
	X_{D} = \{x_j^{D}\}_{j = 1}^{V} = \{o(x_i^{R})\}_{i = 1}^{M}, \forall o \in O,\label{eqn:disturbing}
\end{equation}
where $V \gg M$, creating an expanded dataset of augmented samples $\{x_j^{D}\}_{j = 1}^{V}$ comprising modified versions of the original real-world images. We use this augmented dataset for an intermediate fine-tuning of $G_{R}$'s encoder by minimizing:
\begin{equation}
	\begin{split}
		\mathcal{L}_{G_R}(\varphi; x_i^{D}) = & -\mathcal{D}_{KL}(q_\varphi^{R}(z_i \vert x_i^{D})\|p_\theta(z_i))\\ & + \mathbb{E}_{q_\varphi^{R}(z_i \vert x_i^{D})}[\log p_\theta(x_i^{PS} \vert z_i)].\label{eqn:transformation}
	\end{split}
\end{equation}

For our second technique, we synthesize novel training samples through latent space manipulation. We first train an additional, temporary convolutional VAE, denoted $G_{E}$, exclusively on the original real-world images $\{x_i^{R}\}_{i = 1}^{M}$. This training process yields an encoder $q_\phi^{E}(z_i \vert x_i^{R})$ and decoder $p_\theta(x_i^{R} \vert z_i)$.

We use the encoder of $G_{E}$ to map $X_{R}$ into a latent space. We then apply a particular function $r_{n}^{v}$ to replace values in $n$ random elements of these latent vectors with random values $v$ from a narrow continuous range $[l, u]$. The modified latent vectors, $\tilde{Z} = \{\tilde{z}_j\}_{j = 1}^{W}$, are then passed through the decoder of $G_E$ to generate a new dataset $X_{E}$ of synthesized images:
\begin{equation}
	X_{E} = \{x_j^{E}\}_{j = 1}^{W} = \{\tilde{g}_{G_E}(r_{n}^{v}(z_i))\}_{i = 1}^{M}, 1 \leqslant n \leqslant K, \forall v \in [l, u],\label{eqn:decoding}
\end{equation}
where $X_{E}$ contains significantly more samples than $X_{R}$, since $W \gg M$. This dataset provides novel yet plausible variations of the real-world data. We use $X_{E}$ for final fine-tuning of $G_{R}$'s encoder by minimizing:
\begin{equation}
	\begin{split}
		\mathcal{L}_{G_R}(\varphi; x_i^{E}) = & -\mathcal{D}_{KL}(q_\varphi^{R}(z_i \vert x_i^{E})\|p_\theta(z_i))\\ & + \mathbb{E}_{q_\varphi^{R}(z_i \vert x_i^{E})}[\log p_\theta(x_i^{PS} \vert z_i)].\label{eqn:generation}
	\end{split}
\end{equation}

Together, these augmentation and fine-tuning stages progressively refine the encoder network $q_\varphi^{R}(z_i \vert x_i^{R})$, improving its ability to map varied real-world scenes into the shared latent representation space and thereby enhancing the overall performance of the sim-to-real transfer.

\begin{algorithm*}[h!]
	\caption{Unsupervised training method used by the proposed TL approach.}\label{alg:training}
	\begin{algorithmic}[1]
		
		\Require{
			\\
			Set of simulation images $X_{S} = \{x_i^{S}\}_{i = 1}^{N}$,\\
			Set of real-world images $X_{R} = \{x_i^{R}\}_{i = 1}^{M}$,\\
			Model of synthetic convolutional variational autoencoder $G_{S}$,\\ 
			Model of pseudo-synthetic convolutional variational autoencoder $G_{R}$,\\
			Number $T$ of training/fine-tuning steps.\\
		}
		\Ensure{
			\\
			Common representation space $X_{PS}$ generated from simulation and reality.\\
		}
		
		\For{$t = 1$,...,$T$}
			\State Sample latent variables with $\{z_i\}_{i = 1}^{N} = \{g_{G_S}(x_i^{S})\}_{i = 1}^{N}$ and Eq. (\ref{eqn:sampling}).
			\State Generate synthetic images with $\{\tilde{x}_i^{S}\}_{i = 1}^{N} = \{\tilde{g}_{G_S}(z_i)\}_{i = 1}^{N}$.
			\State Update the weight parameters $\phi$ and $\theta$ of $G_{S}$ with loss in Eq. (\ref{eqn:loss}).
		\EndFor
		
		\\
		\State Load into $G_{R}$ the parameters $\phi$ and $\theta$ learned for $G_{S}$, and freeze the decoder of $G_{R}$.\\
		
		\For{$t = 1$,...,$T$}
			\State Sample latent variables with $\{z_i\}_{i = 1}^{M} = \{g_{G_R}(x_i^{R})\}_{i = 1}^{M}$ and Eq. (\ref{eqn:sampling}).
			\State Generate pseudo-synthetic images with $\{\tilde{x}_i^{PS}\}_{i = 1}^{M} = \{\tilde{g}_{G_R}(z_i)\}_{i = 1}^{M}$.
			\State Update the weight parameters $\varphi$ of $G_{R}$ with loss in Eq. (\ref{eqn:loss}).
		\EndFor
		
		\\
		\State Produce realistic transformed data $X_{D} = \{x_j^{D}\}_{j = 1}^{V}$ with Eq. (\ref{eqn:disturbing}).\\
		
		\For{$t = 1$,...,$T$}
			\State Sample latent variables with $\{z_j\}_{j = 1}^{V} = \{g_{G_R}(x_j^{D})\}_{j = 1}^{V}$ and Eq. (\ref{eqn:sampling}).
			\State Generate pseudo-synthetic images with $\{\tilde{x}_j^{PS}\}_{j = 1}^{V} = \{\tilde{g}_{G_R}(z_j)\}_{j = 1}^{V}$.
			\State Update the weight parameters $\varphi$ of $G_{R}$ with loss in Eq. (\ref{eqn:loss}).
		\EndFor
		
		\\
		\State Build an extra model $G_{E}$ and train it on real-world images $X_{R} = \{x_i^{R}\}_{i = 1}^{M}$.
		\State Produce realistic generated data $X_{E} = \{x_j^{E}\}_{j = 1}^{W}$ with $G_{E}$ and Eq. (\ref{eqn:decoding}).\\
		
		\For{$t = 1$,...,$T$}
			\State Sample latent variables with $\{z_j\}_{j = 1}^{W} = \{g_{G_R}(x_j^{E})\}_{j = 1}^{W}$ and Eq. (\ref{eqn:sampling}).
			\State Generate pseudo-synthetic images with $\{\tilde{x}_j^{PS}\}_{j = 1}^{W} = \{\tilde{g}_{G_R}(z_j)\}_{j = 1}^{W}$.
			\State Update the weight parameters $\varphi$ of $G_{R}$ with loss in Eq. (\ref{eqn:loss}).
		\EndFor
		
	\end{algorithmic}
\end{algorithm*}

\section{Experiments}\label{sec:experiments}

This section presents a comprehensive empirical evaluation of our proposed hybrid transfer learning framework, designed to validate the three core hypotheses articulated in Section \ref{sec:introduction}. We conducted a series of experiments to assess the efficacy of our approach, which synergizes domain randomization and domain adaptation to bridge the simulation-to-reality gap. The evaluation is structured around three balanced, labeled test datasets comprising real images previously unseen during training and three distinct deep neural network architectures, each designed to assess a specific learning paradigm:

\begin{enumerate}[1.]
	\renewcommand{\labelenumi}{(\arabic{enumi})}
	
	\item A baseline model trained and evaluated exclusively on real image data, representing a conventional real-world-only training approach.
	
	\item A deep neural network architecture trained entirely on synthetic data via the static DR strategy detailed in Section \ref{sec:methodology}, and subsequently evaluated on real test image data to quantify the extent of the reality gap.
	
	\item Our proposed pseudo-synthetic architecture, which is first pre-trained on the extensive simulation dataset and then fine-tuned using the limited real-world dataset via feature-level domain adaptation. The model's alignment between domains is further enhanced through the application of the two data augmentation techniques described in Subsection \ref{subsec:augmentation}. This final architecture is also evaluated on real test image data.
	
\end{enumerate}

The subsequent analysis encompasses a qualitative assessment of the generative models' capabilities, a quantitative evaluation of the learned classification policies, and an in-depth analysis of the resulting latent representation spaces. To validate the method's suitability for resource-constrained platforms, we include an estimation of its computational efficiency. Finally, we present a proof-of-concept case study detailing the successful deployment of the proposed policy on a low-cost physical robot performing a reactive, collision-free exploration task in a real-world environment.

\subsection{Experimental Protocols and Architectures}\label{subsec:architectures}

This subsection details the specific implementation of the training and evaluation protocols for each of the three experimental paradigms. We describe the deep learning architectures employed to learn three distinct image classification policies, each applicable to vision-based robotic navigation in real-world environments.

In Experiment (1), which serves as a real-world baseline, a convolutional VAE was trained using unsupervised learning [\citenum{shanthamallu_brief_2017}] exclusively on the 4556 real-world images. This model's primary function was to perform unsupervised feature extraction, generating abstract, low-dimensional representations from a small, labeled subset of 186 real images, previously preprocessed to 64x64 RGB format. These compact 32-element latent vectors subsequently served as input for training a lightweight Multi-Layer Perceptron (MLP) [\citenum{pierson_deep_2017}] via supervised learning [\citenum{shanthamallu_brief_2017}]. The classifier's simple architecture, comprising a single hidden layer of 100 neurons and an output layer of 4 neurons, was intentionally designed for memory efficiency. Given the inherent challenges of acquiring a large and balanced labeled dataset from reality, we employed a stratified 10-fold cross-validation [\citenum{mahesh_stratified_2023}] methodology to ensure a robust and reliable evaluation of the resulting policy.

For Experiment (2), representing a pure domain randomization methodology, a convolutional VAE was trained with unsupervised learning on the large-scale dataset of 45225 synthetic images. This process, which implements our procedural randomization strategy by exposing the model to a wide variety of simulated conditions, aimed to learn a policy with a high degree of generalization. The VAE's encoder generated low-dimensional representations from the synthetic data, which were then used to train a supervised classification network with a structure identical to that of the baseline experiment. This synthetic architecture was subsequently evaluated on real test scenarios to quantitatively assess the performance of a direct sim-to-real policy transfer without any adaptation to the target domain.

In Experiment (3), our proposed hybrid transfer learning method was evaluated. Its core is a dual VAE architecture employing a shared decoder, a structure inspired by the work of Inoue \textit{et al.} [\citenum{inoue_transfer_2018}]. However, our method significantly advances this concept by applying it to a much larger and more challenging dataset: 45225 synthetic and 4556 real images, compared to 4131 synthetic and only 54 paired real images in their work. Additionally, our TL method integrates two complementary data augmentation techniques to further enhance domain alignment. This scaled-up approach facilitates a more robust alignment of the latent space, enabling the high performance reported in our findings.

The process begins by taking the pre-trained variational autoencoder from the DR stage (\textit{i.e.}, Experiment (2)). We then perform domain adaptation by fine-tuning its encoder using the 4556 unlabeled real-world images, while the parameters of the shared decoder remain frozen. To augment the limited real-world data and improve the robustness of this adaptation, we first generated an expanded balanced set of 34600 samples through standard image-space transformations (\textit{e.g.}, cropping, brightness, contrast, and saturation adjustments). Secondly, we synthesized 18500 novel samples by systematically perturbing the latent space representations of the real-world images. This was achieved by slightly modifying the values of only 2 random elements out of the 32 that constituted each latent vector.

The final pseudo-synthetic policy was obtained through a sequential, two-stage training process for the MLP classifier. Initially, the classification neural network was trained on the latent vectors derived from the 45225 synthetic images to learn a generalized base policy. Subsequently, this model was fine-tuned using the 18500 compressed representations corresponding to the augmented samples generated via latent-space manipulation, thereby allowing the final policy to benefit from both broad generalization and targeted adaptation.

\subsection{Training and Evaluation Datasets}\label{subsec:datasets}

\begin{figure*}[h!]
	\centering
	\includegraphics[width=\textwidth]{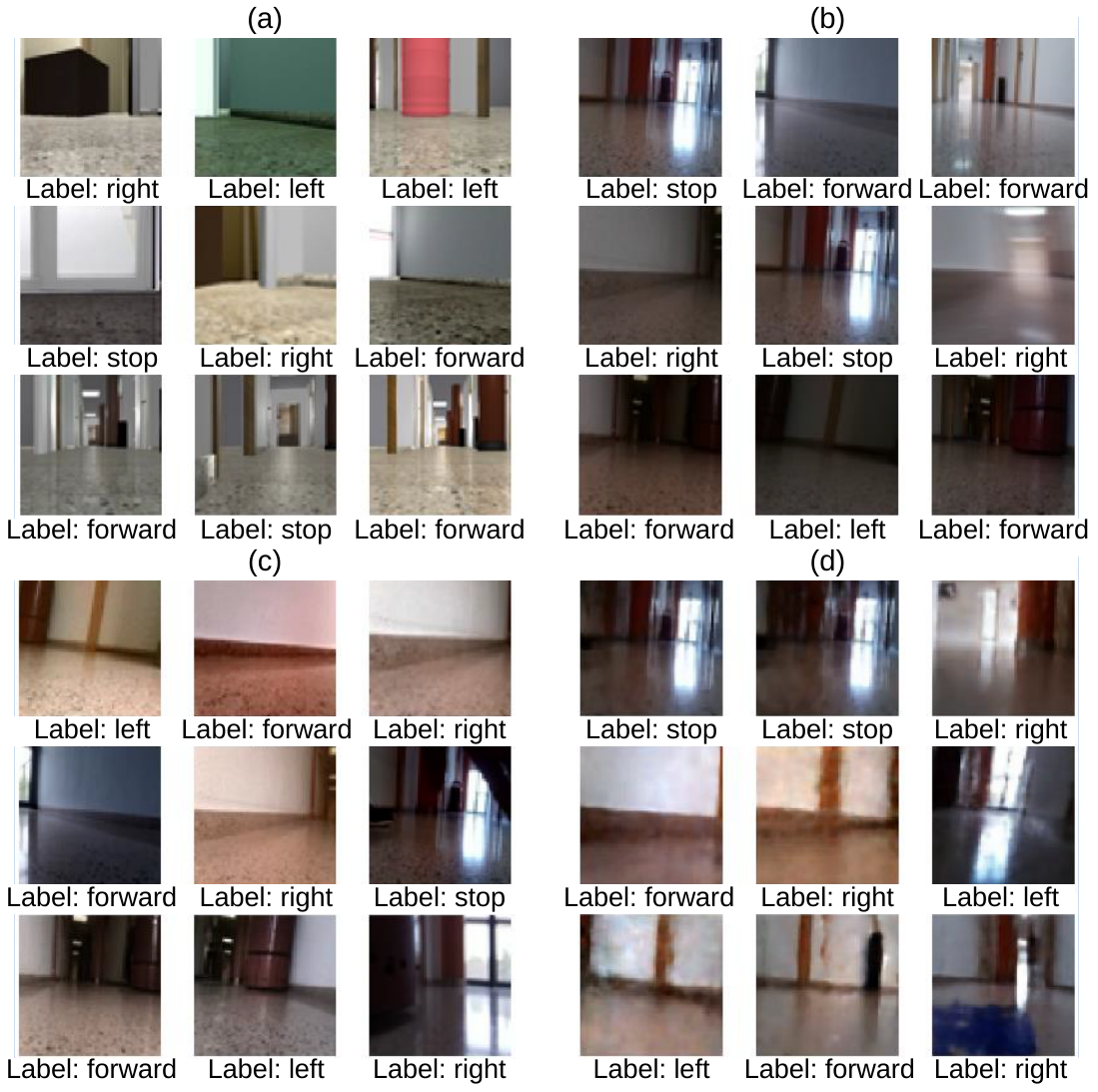}
	\caption{Examples of labeled images used across our three experiments for training, fine-tuning, and testing. (a) Synthetic images from the simulated environment, used for the domain randomization training stage in Experiments (2) and (3). (b) (b) Real-world images captured in the physical environment, used for baseline training in Experiment (1), fine-tuning in Experiment (3), and evaluation in all experiments. (c) Realistically transformed images used to augment the real-world dataset for fine-tuning in Experiment (3) and evaluation in all experiments. (d) Synthesized realistic images, generated via latent-space manipulation, used to further refine the pseudo-synthetic autoencoder in Experiment (3).}\label{fig:datasets}
\end{figure*}

The empirical validation of our transfer learning method relies on several distinct datasets for training and evaluation, sourced from both the simulated and real-world domains. As detailed in Subsection \ref{subsec:acquisition} our data acquisition protocol yielded two primary training sets. The first is an extensive, balanced collection of 45225 synthetic images captured within our high-fidelity physics simulator, which serves as the foundation for the domain randomization stage. The second is a much smaller set of 4556 labeled images captured with the physical robot in the corresponding real-world environment, intended for the domain adaptation phase. To facilitate supervised learning, all samples across both datasets were annotated with one of four possible navigation commands: \textit{forward}, \textit{right}, \textit{left}, and \textit{stop}, as shown in Fig. \ref{fig:datasets}(a) and Fig. \ref{fig:datasets}(b).

To enhance the domain adaptation process and further bridge the sim-to-real gap, the limited real-world dataset was expanded using the two complementary data augmentation techniques presented in Subsection \ref{subsec:augmentation}. This process generated additional realistic training samples through both image-space transformations and latent-space manipulation. Fig. \ref{fig:datasets}(c) and Fig. \ref{fig:datasets}(d) provide illustrative examples from each of these data categories.

For a rigorous and unbiased assessment of the learned policies, we curated three distinct, balanced, and labeled test datasets, all of which were entirely unseen during any training or fine-tuning phase. The composition of these evaluation sets is detailed in Table \ref{tab:datasets}. The first set comprises non-augmented real-world images and was created using a 10-fold cross-validation protocol. The other two, larger sets consist of realistically transformed images, produced by applying slight, random modifications (\textit{e.g.}, to brightness, contrast, and saturation) to the original real-world samples. These augmented sets were designed to systematically test the generalization and adaptation capabilities of the learned policies under plausible environmental variations that a robot would encounter in practice. For instance, random modifications to image brightness help simulate different lighting conditions that might occur in real-world settings.

\begin{table}[h!]
	\begin{minipage}{\columnwidth}
		\begin{center}
			\caption{Evaluation datasets comprising real images previously unseen during the training and fine-tuning phases.}\label{tab:datasets}
			\begin{tabular}{@{}ccc@{}}
				\hline
				\textbf{Identifier} & \textbf{Name} & \textbf{Size}\\
				\hline
				\hline
				(CV) & Cross-Validation Real-World Test & 186\\
				(T1) & Realistic Transformations Test 1 & 3400\\
				(T2) & Realistic Transformations Test 2 & 4000\\
				\hline
			\end{tabular}
		\end{center}
	\end{minipage}
\end{table}

\subsection{Training Configuration and Optimizer Selection}\label{subsec:setup}

This subsection outlines the configuration of the training phase for our deep neural network architectures. The core of the proposed training methodology, detailed in Algorithm \ref{alg:training}, is centered on the variational autoencoders, as the subsequent classification neural network is a very compact multi-layer perceptron. As specified in Subsection \ref{subsec:architectures}, our image classification model consists of only three layers. Therefore, the primary challenge lies in effectively training and, most critically, fine-tuning our autoencoders to successfully bridge the simulation-to-reality gap.

To achieve stable and efficient convergence, we employed the Root Mean Square Propagation (RMSProp) optimizer. RMSProp is an adaptive learning rate method developed to overcome some limitations of standard stochastic gradient descent by adapting the step size using mean moments. Its mechanism limits oscillations during the descent phase, allowing for more substantial and effective steps in the correct direction of convergence [\citenum{abdulkadirov_survey_2023}]. These adaptive capabilities are highly advantageous for a deep transfer learning problem such as this, particularly during the delicate fine-tuning stage where the model must adapt to the subtle visual discrepancies of the real domain (\textit{e.g.}, textures, lighting, and sensor noise) without catastrophically forgetting the robust, general features learned from simulation.

The specific hyperparameter values used for both the initial training and the subsequent fine-tuning stages are presented in Table \ref{tab:hyperparameters}. The choice of a conservative learning rate, combined with the adaptive nature of RMSProp, provides the necessary stability for the fine-tuning process. This combination constitutes a robust training protocol, enabling our pseudo-synthetic VAE to effectively learn a powerful common representation space for successful sim-to-real transfer.

\begin{table}[h!]
	\begin{minipage}{\columnwidth}
		\begin{center}
			\caption{Key hyperparameter values configured for the training and subsequent fine-tuning stages of the variational autoencoder models.}\label{tab:hyperparameters}
			\begin{tabular}{@{}ccc@{}}
				\hline
				\textbf{Symbol} & \textbf{Description} & \textbf{Value}\\
				\hline
				\hline
				$\alpha$ & Learning rate & 0.001\\
				$B$ & Batch size & 32\\
				$T$ & Training/fine-tuning steps & 200\\
				\hline
			\end{tabular}
		\end{center}
	\end{minipage}
\end{table}

\subsection{Qualitative Analysis of Reconstruction Capabilities}\label{subsec:reconstruction}

This subsection presents a qualitative assessment of the deep generative models trained across our three experiments. The evaluation focuses on the image reconstruction quality of each convolutional variational autoencoder, which was tasked with generating images from real input samples randomly selected from the test datasets detailed in Table \ref{tab:datasets}. The process involved encoding the 64x64 RGB test images into 32-dimensional latent vectors, which then served as input to the respective decoders for reconstruction.

It is crucial to frame this analysis within the specific goals of our application. The primary objective of our variational autoencoders is not to achieve photo-realistic, high-fidelity reconstructions, a task for which other specialized generative architectures are better suited [\citenum{razavi_generating_2019, tewari_high-fidelity_2020}]. Instead, for tasks like autonomous navigation, the characteristically blurred output of VAEs can be re-contextualized as a powerful form of implicit regularization.

The process of compressing high-dimensional visual information through a low-dimensional latent bottleneck naturally filters out high-frequency, domain-specific details (\textit{e.g.}, wall textures). Simultaneously, it preserves the low-frequency, structural information (\textit{e.g.}, corridor geometry and obstacle shapes) that is most critical for embodied navigation and most likely to be invariant across the sim-to-real gap. This capacity for feature abstraction is particularly effective for low-cost robotic platforms that rely on a single monocular RGB camera, as it facilitates the extraction of task-relevant cues from limited sensory input, enabling efficient and rapid decision-making for real-time reactive navigation.

\begin{figure*}[h!]
	\centering
	\includegraphics[width=\textwidth]{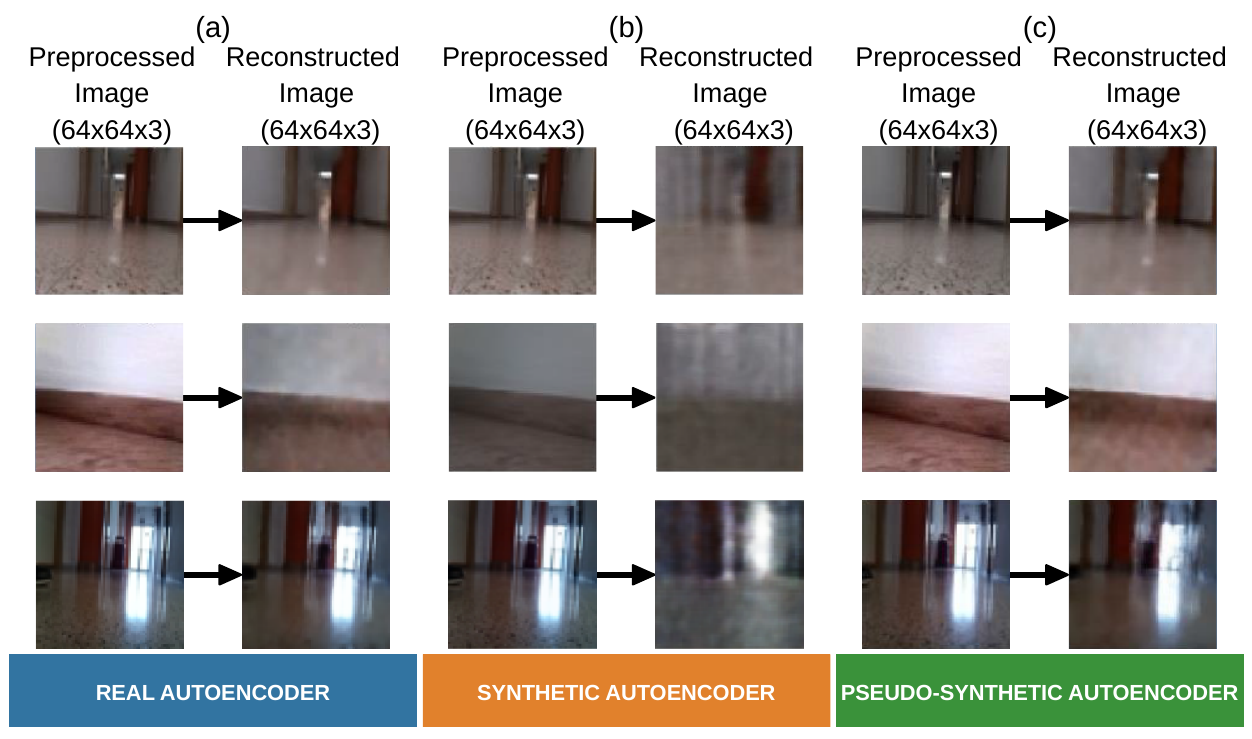}
	\caption{Reconstructions generated by the decoder of our three convolutional variational autoencoders from input real image data. (a) Test images reconstructed using the real-only autoencoder trained for Experiment (1). (b) Test images reconstructed through the sim-only autoencoder trained in Experiment (2). (c) Test images reconstructed using our proposed pseudo-synthetic autoencoder trained and fine-tuned for Experiment (3).}\label{fig:reconstructions}
\end{figure*}

Our observations confirmed a direct correlation: deep generative models that produced higher-quality structural reconstructions yielded better downstream classification performance. As shown in Fig. \ref{fig:reconstructions}(a) and Fig. \ref{fig:reconstructions}(c), the VAEs from Experiments (1) and (3) successfully reconstructed the global structure of our real indoor environment, accurately capturing the obstacles and general lighting conditions present in the original images. These two DL models, which produced reconstructions of similar quality, exhibited some difficulty only in reconstructing high-frequency details, such as the floor texture, which aligns with the expected behavior of the variational autoencoder as a structural feature extractor.

In contrast, and as anticipated, the generative model from Experiment (2) produced more blurred reconstructions, a direct consequence of its exclusive training on simulation data. However, even in this case, as depicted in Fig. \ref{fig:reconstructions}(b), the global structure and key obstacles within the real environment remained identifiable, demonstrating that the model still learned geometrically salient features despite the significant domain shift.

\subsection{Quantitative Analysis of Classification Performance}\label{subsec:classification}

To complement the qualitative findings, this subsection presents a quantitative assessment of the classification policies learned across our three experimental setups. We evaluated the performance of each classification neural network using the autoencoders previously analyzed. The evaluation was conducted on the three balanced, labeled test datasets detailed in Table \ref{tab:datasets}, calculating mean accuracy rates along with corresponding median and variance values.

\begin{figure*}[h!]
	\centering
	\includegraphics[width=\textwidth]{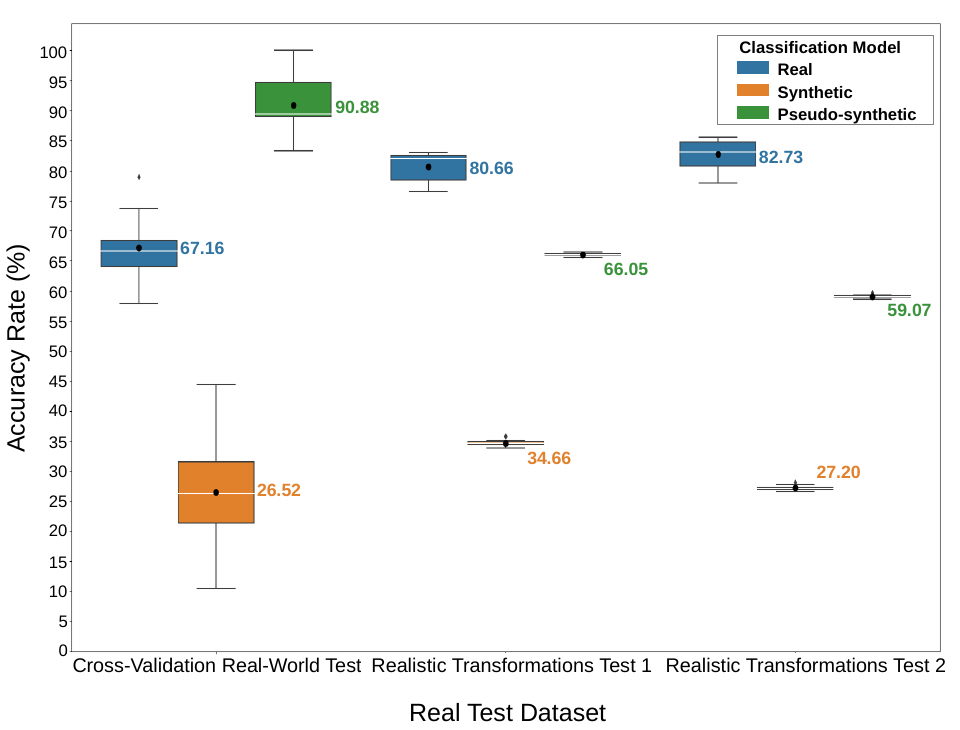}
	\caption{Mean accuracy rates, along with corresponding median and variance values, obtained when predicting probabilities related to the four possible ground-truth labels used to annotate input real images. We achieved the final results using three distinct classification models: the real-only image classifier (blue) trained for Experiment (1), the sim-only image classifier (orange) trained in Experiment (2), and our proposed pseudo-synthetic image classifier (green) trained and fine-tuned for Experiment (3). The reported accuracy percentages resulted from evaluating each classifier using three different test datasets, presented in decreasing order of importance.}\label{fig:probabilities}
\end{figure*}

These metrics were obtained when using our three distinct deep neural network architectures to predict probabilities associated with the four navigation categories (\textit{i.e.}, \textit{forward}, \textit{right}, \textit{left}, and \textit{stop}). To ensure statistical robustness, the policy performance on the non-augmented real-world test set was measured using a 10-fold cross-validation protocol, while results on the two realistically transformed test sets were averaged over 10 independent runs. The results are summarized in Fig. \ref{fig:probabilities}.

The real-only model from Experiment (1) achieved a moderate mean accuracy of 67.16\% on the primary (CV) test set. This performance reflects the inherent limitations of training a deep learning model with a small, sparsely-distributed real-world dataset. However, this deep neural network architecture showed remarkable adaptability to plausible environmental variations, achieving mean accuracy rates exceeding 80\% for the two sets of realistic samples generated through image-space transformations.

Conversely, the synthetic model from Experiment (2), trained exclusively on simulated data, performed poorly, failing to surpass a 35\% mean accuracy on any real test set. This result quantitatively underscores the significant reality gap that persists when domain adaptation is not employed.

In contrast, our proposed pseudo-synthetic model from Experiment (3) demonstrated superior predictive capacity, achieving a mean accuracy of 90.88\% on the (CV) test set. However, compared to Experiment (1), this deep neural network architecture showed lower adaptation ability to the varying circumstances represented in our last two test sets, achieving mean accuracy rates of approximately 60\%.

The significant improvement achieved by the pseudo-synthetic model on the (CV) test dataset highlights the efficacy of our hybrid transfer learning strategy, centered on the dual encoder, shared decoder architecture. While a direct ablation study comparing shared versus independent decoders was not conducted, this architectural choice is rigorously supported by established principles in domain adaptation and incremental learning.

Crucially, our design adheres to the \textquotedblleft{shared latent space assumption}\textquotedblright\text{ }formally analyzed by Huang \textit{et al.} [\citenum{huang_multimodal_2018}], who posit that corresponding images in different domains can be mapped to a common latent code. In a dual VAE framework, the shared decoder acts as a critical structural constraint that forces the latent spaces of both simulation and reality to align within the same semantic manifold. Without this shared component, there is no mathematical pressure to ensure that a latent representation $z_i$ carries the same navigation-relevant meaning across domains, leading to a divergence that would render the simulation-trained policy ineffective in real-world scenarios.

Furthermore, the use of a shared decoder is a recognized mechanism for mitigating catastrophic forgetting during the fine-tuning of deep generative models. As demonstrated by Dai \textit{et al.} [\citenum{dai_multi-task_2020}] in the context of multi-task incremental learning, unshared decoders often result in the overwriting of source knowledge because gradients derived from the target domain only update domain-specific paths, causing a sudden decrease in performance for the original tasks. Their empirical results showed a significant performance gap when shifting from shared to unshared decoders (\textit{e.g.}, a drop in category extraction for sentiment analysis from 0.913 to 0.842), validating that parameter sharing weakens the interference between old and new knowledge.

In our specific case of low-cost robotic navigation, where real-world data is limited to only 4556 images, the shared decoder serves as a \textquotedblleft{generative constraint}\textquotedblright\text{ }[\citenum{yang_generalizing_2023}]. This constraint amortizes the structural knowledge learned from our 45225 simulated images, acting as a regularizer that prevents the encoder from overfitting to the small real-world dataset. By freezing the decoder parameters during adaptation, a technique also explored by Inoue \textit{et al.} [\citenum{inoue_transfer_2018}], we ensure that the latent space remains anchored to the robust geometry learned in simulation while the encoder learns to map noisy physical inputs into this pre-structured space. Consequently, the shared decoder is not merely an empirical heuristic but a systemic requirement for stable sim-to-real transfer under resource-constrained conditions, leading to the high 90.88\% classification performance observed.

\subsection{Ablation Study: Impact of Data Augmentation Strategy}
\label{subsec:ablation}

To rigorously quantify the contribution of the domain alignment strategies detailed in Subsection \ref{subsec:augmentation}, and to distinguish the efficacy of our full hybrid framework from a standard dual VAE architecture (\textit{e.g.}, as proposed in related works such as Inoue \textit{et al.} [\citenum{inoue_transfer_2018}]), we conducted an ablation study. In this experiment, we trained the exact same neural network architecture described in Subsection \ref{subsec:architectures} using the configuration of Experiment (3), but with the specific exclusion of the two complementary data augmentation stages (\textit{i.e.}, image-space transformations and latent-space manipulation). Consequently, the model relied solely on the shared decoder constraint to align the 45225 simulated images and the 4556 real-world images without the expansive bridging provided by the augmented datasets.

The results, evaluated on our 4-action space (\textit{i.e.}, move forward, turn right, turn left, and stop) across the three test datasets, are presented in Table \ref{tab:ablation}. The comparison reveals a critical dependency on the proposed augmentation techniques. When the augmentation strategies are removed, the mean accuracy on the primary (CV) dataset precipitates from 90.88\% to 47.06\%. This represents a performance drop of 43.82 percentage points. Similarly, performance on the realistically transformed datasets (T1) and (T2) degrades to levels only marginally superior to the simulation-only baseline (\textit{i.e.}, Experiment (2)), with accuracies of 39.15\% and 34.20\%, respectively.

\begin{table*}[h!]
	\begin{minipage}{\textwidth}
		\begin{center}
			\caption{Ablation study mean accuracy rates comparing the proposed full hybrid framework against a baseline implementation without the data augmentation strategies described in Subsection \ref{subsec:augmentation}.}\label{tab:ablation}
			\begin{tabular}{@{}cccc@{}}
				\hline
				\textbf{Real Test Dataset} & \textbf{Base Dual VAE} & \textbf{Proposed Method} & \textbf{Contribution}\\
				\hline
				\hline
				(CV) & 47.06\% & \textbf{90.88\%} & +43.82\%\\
				(T1) & 39.15\% & \textbf{66.05\%} & +26.90\%\\
				(T2) & 34.20\% & \textbf{59.07\%} & +24.87\%\\
				\hline
			\end{tabular}
		\end{center}
	\end{minipage}
\end{table*}

These findings provide empirical evidence that the dual VAE architecture with a shared decoder is, by itself, insufficient to bridge the reality gap when dealing with limited, unpaired real-world data. While the shared decoder provides a necessary structural constraint, the ablation results demonstrate that the high generalization capability of our system, able to obtain nearly 91\% accuracy, is primarily driven by the novel alignment achieved through the proposed augmentation stages. These techniques are therefore confirmed not as optional enhancements, but as fundamental components required to effectively map the sparse real-world distribution into the shared latent manifold.

\subsection{Policy Analysis for Reactive Collision-Free Navigation}\label{subsec:policy}

To further examine the practical effectiveness of the learned policies, this subsection evaluates their potential application within a reactive, map-less navigation system. The scenario involves a low-cost robot equipped with a single monocular RGB camera exploring a real indoor corridor, making decisions based on the test images from our evaluation datasets.

For this task-oriented analysis, we simplified the classification problem by reducing the action space from four distinct commands to three functional categories: \textit{forward}, \textit{turn}, and \textit{stop}. In this simplified framework, we considered predictions correct if the network selected any turning action (\textit{i.e.}, turn right or turn left) when the ground truth indicated either \textit{right} or \textit{left} turns. This simplification is justified in reactive collision avoidance, where the imperative to initiate an evasive turn to avoid an obstacle is often more critical than the specific direction of the maneuver.

\begin{figure*}[h!]
	\centering
	\includegraphics[width=\textwidth]{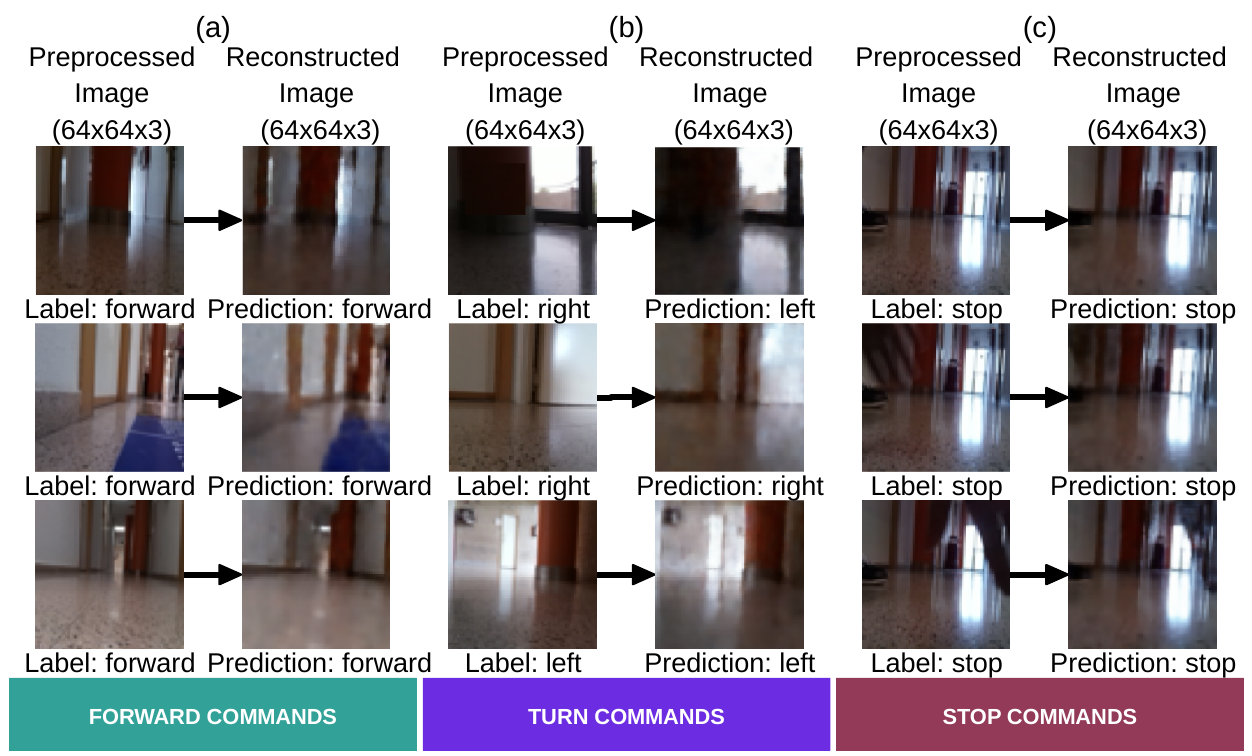}
	\caption{Predictions and reconstructions obtained from labeled real images using our proposed pseudo-synthetic model trained in Experiment (3). These tests focused on reactive collision-free exploration tasks in the real world, involving reconstruction of input images and their classification into three categories: \textit{forward}, \textit{turn}, and \textit{stop}. (a) Test image data representing scenarios requiring forward actions. (b) Test image data corresponding to situations requiring turn commands. (c) Test image data representing circumstances requiring stop actions.}\label{fig:predictions}
\end{figure*}

The analysis of our proposed pseudo-synthetic policy from Experiment (3) demonstrates its high potential for successful real-world deployment. As illustrated in Fig. \ref{fig:predictions}(b), the model exhibits intelligent reactive behavior. For instance, in a scenario consisting of a column beside a window, the ground-truth label indicated a right turn. In this case, the policy still predicted a correct turning action, albeit in the opposite direction (\textit{i.e.}, to the left). In a reactive context, this is a successful outcome, as the robot would still perform the necessary evasive maneuver to avoid collision and continue its exploration.

Furthermore, as shown in Fig. \ref{fig:predictions}(a) and Fig. \ref{fig:predictions}(c), our transfer learning approach would also enable successful collision-free exploration in scenarios requiring forward movement in open spaces or stopping. Notably, the robot would successfully execute three consecutive stop commands to avoid collision with a person standing directly in front of it while reaching toward its camera.

As expected, evaluating the learned policies under this simplified three-category framework led to a general improvement in their mean accuracy rates. The updated quantitative results for all three architectures are presented in Table \ref{tab:probabilities}. Notably, our pseudo-synthetic policy maintained its superior accuracy of 90.88\% on the primary (CV) test set, reinforcing the robustness of our proposed transfer learning approach for practical navigation tasks.

\begin{table*}[h!]
	\begin{minipage}{\textwidth}
		\begin{center}
			\caption{Updated mean accuracy rates for the three models evaluated on the simplified three-category classification task.}\label{tab:probabilities}
			\begin{tabular}{@{}ccc@{}}
				\hline
				\textbf{Real Test Dataset} & \textbf{Experiment} & \textbf{Mean Accuracy Rate (\%)}\\
				\hline
				\hline
				(CV) & (1) & 72.57\\
				(CV) & (2) & 32.92\\
				(CV) & (3) & \textbf{90.88}\\
				\hline
				(T1) & (1) & \textbf{81.76}\\
				(T1) & (2) & 40.85\\
				(T1) & (3) & 69.86\\
				\hline
				(T2) & (1) & \textbf{83.91}\\
				(T2) & (2) & 35.04\\
				(T2) & (3) & 60.78\\
				\hline
			\end{tabular}
		\end{center}
	\end{minipage}
\end{table*}

\subsection{Analysis of Latent Representation Spaces}\label{subsec:clustering}

\subsubsection{Qualitative Evaluation of Latent Space Organization}\label{subsubsec:tsne}

A fundamental challenge in autonomous navigation is the ability to learn a well-structured latent space that effectively encodes meaningful spatial features from raw visual inputs. This subsection examines the latent spaces generated by the three convolutional variational autoencoders from our experiments. This analysis serves as a practical and insightful alternative to direct real-world deployment, offering valuable information about the quality of the learned representations and the expected system behavior. For this purpose, we utilize the simplified three-category action space (\textit{forward}, \textit{turn}, and \textit{stop}) established in the preceding subsection.

\begin{figure*}[h!]
	\centering
	\includegraphics[height=0.88\textheight, width=0.65\textwidth]{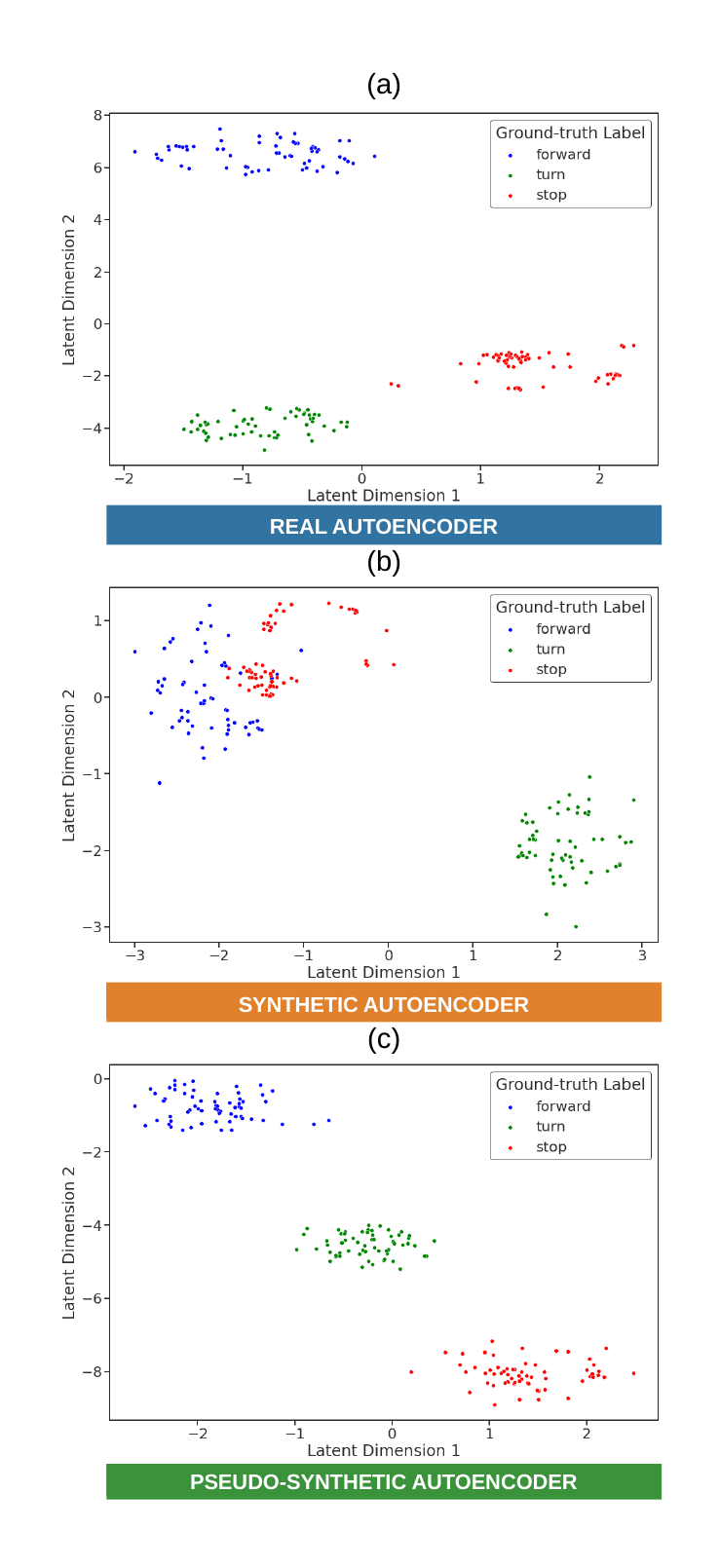}
	\caption{Two-dimensional t-SNE visualization of the latent spaces generated from the non-augmented real-world (CV) test set. (a) Latent space from the real-only autoencoder. (b) Latent space from the sim-only autoencoder. (c) Latent space from the proposed pseudo-synthetic autoencoder.}\label{fig:cv}
\end{figure*}

To visualize and interpret the 32-dimensional latent vectors produced by our autoencoders, we employed the t-Distributed Stochastic Neighbor Embedding (t-SNE) technique. This non-linear dimensionality reduction method is particularly effective for visualizing high-dimensional data, as it excels at preserving local relationships between data points and revealing the underlying cluster structure in a low-dimensional map [\citenum{maaten_visualizing_2008}]. By mapping high-dimensional representations to a 2D space, t-SNE allows for the identification of clusters corresponding to different navigation commands, thus revealing the organizational quality of the learned latent space.

The t-SNE visualizations reveal different organizational patterns across the three models. For the non-augmented real-world (CV) test set, as observed in Fig. \ref{fig:cv}(c), the pseudo-synthetic autoencoder produced a remarkably well-organized latent space with clearly separated clusters corresponding to the different navigation actions. This highly structured representation directly correlates with its superior mean classification accuracy of 90.88\%. In contrast, as shown in Fig. \ref{fig:cv}(a), the real-only model produced a space with moderately defined clusters, consistent with its 72.57\% accuracy. The sim-only autoencoder, in turn, generated a largely unstructured space with significant class overlap, depicted in Fig. \ref{fig:cv}(b), reflecting its poor generalization and 32.92\% accuracy.

\begin{figure*}[h!]
	\centering
	\includegraphics[height=0.88\textheight, width=0.65\textwidth]{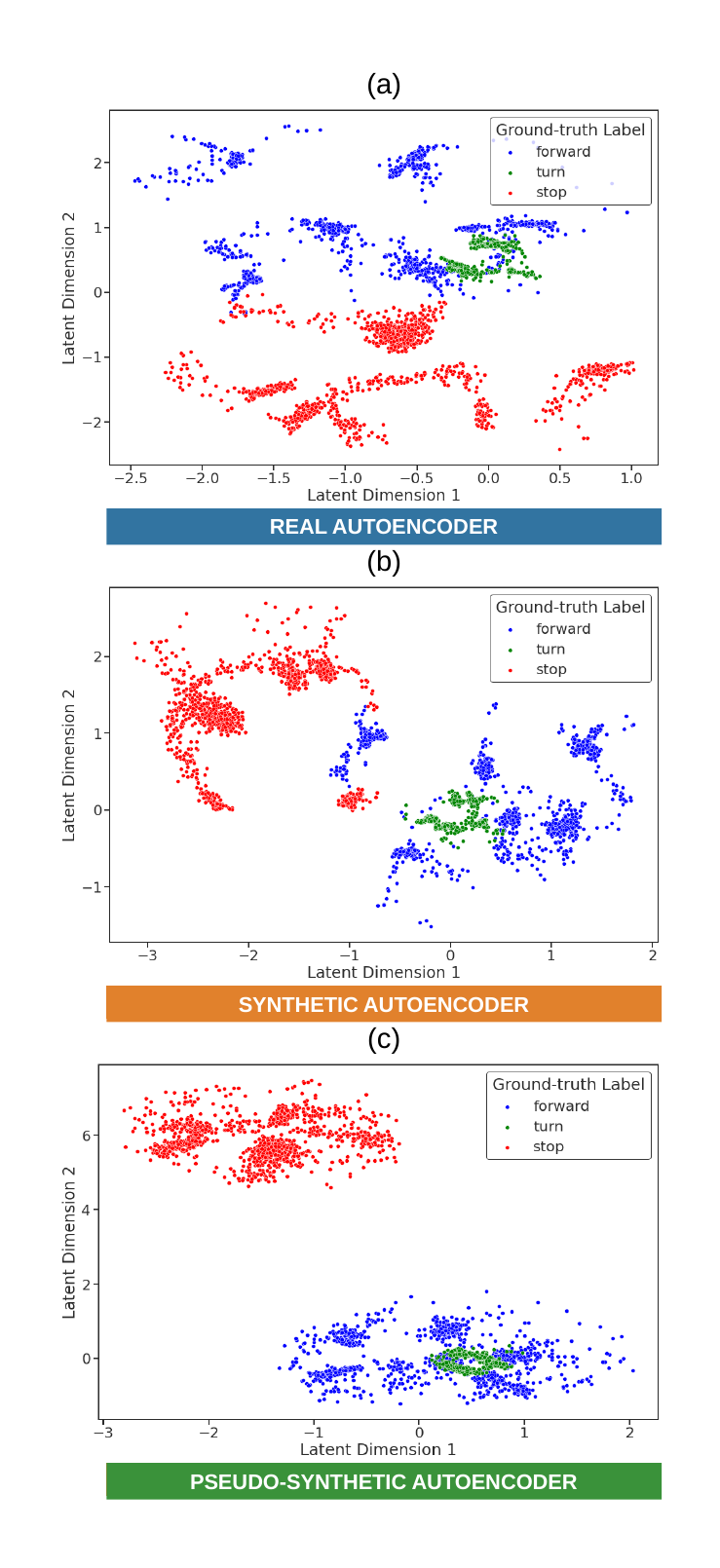}
	\caption{Two-dimensional t-SNE visualization of the latent spaces generated from the realistically transformed (T1) test set. (a) Latent space from the real-only autoencoder. (b) Latent space from the sim-only autoencoder. (c) Latent space from the proposed pseudo-synthetic autoencoder.}\label{fig:t1}
\end{figure*}

\begin{figure*}[h!]
	\centering
	\includegraphics[height=0.88\textheight, width=0.65\textwidth]{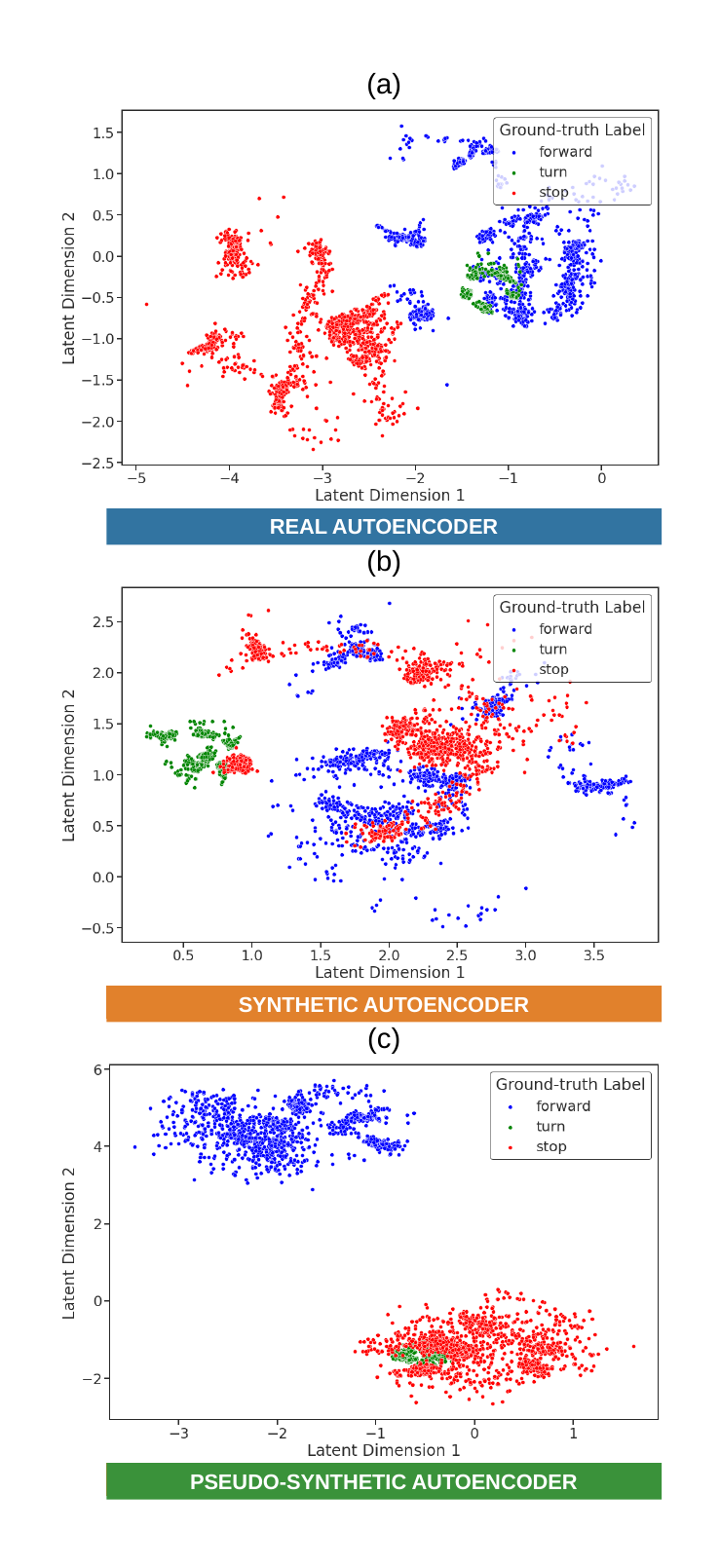}
	\caption{Two-dimensional t-SNE visualization of the latent spaces generated from the realistically transformed (T2) test set. (a) Latent space from the real-only autoencoder. (b) Latent space from the sim-only autoencoder. (c) Latent space from the proposed pseudo-synthetic autoencoder.}\label{fig:t2}
\end{figure*}

For the augmented test sets (T1) and (T2), depicted in Fig. \ref{fig:t1} and Fig. \ref{fig:t2}, the latent spaces of the real-only autoencoder demonstrated improved cluster formation, aligning with enhanced mean classification accuracies of 81.76\% and 83.91\%, respectively. In contrast, poor organizational patterns persisted when using the sim-only model, especially with the (T2) dataset. Finally, the latent spaces of the pseudo-synthetic autoencoder showed somewhat reduced cluster definition compared to the (CV) set, corresponding to its lower classification performance on these specifically transformed datasets.

The organization of samples within the latent spaces provides significant information about the capability of our proposed VAE to learn visual features relevant for autonomous reactive navigation. Several observable subclusters within the three action categories in the pseudo-synthetic latent spaces likely represent variations within specific navigation actions (\textit{e.g.}, distinguishing between right and left turns, or between forward movements in different areas of the environment). These detailed distinctions might contribute significantly to the superior performance of the model in real-world scenarios.

The relative positions of clusters within the latent space also reveal specific relationships between different navigation actions. For instance, the spatial proximity between forward and turn clusters in Fig. \ref{fig:t1}(c) might indicate a learned understanding of the sequential relationship between these two action categories, particularly appropriate for autonomous exploration applications.

In turn, the stop action usually formed a more clearly separated cluster, reflecting its fundamental difference from actions involving movement. Nevertheless, as depicted in Fig. \ref{fig:t2}(c), there is significant overlap between the categories of turning and stopping actions. This could lead to errors in subsequent classification of input samples and, consequently, to a considerable decrease in system performance. In a potential exploration application, this limitation could translate into suboptimal decisions. For instance, the robot might stop unnecessarily in a space where a turn action would provide greater environmental coverage. Conversely, it could lead to critical errors, such as attempting to turn when facing a close obstacle where only a stop action could prevent a collision.

\subsubsection{Quantitative Evaluation of Cluster Validity}\label{subsubsec:cvi}
	
To complement this visual assessment and provide a more rigorous evaluation, we conducted a quantitative analysis of the generated latent spaces. This analysis employs three widely-recognized internal Cluster Validity Indices (CVIs) [\citenum{ikotun_cluster_2025}] to measure the quality of the clusters formed in the t-SNE representations. The selected metrics are:

\begin{itemize}
	
	\item The Silhouette Index, which evaluates how well-separated and compact clusters are. Its value ranges from -1 to 1, where a score closer to 1 indicates that data points are well-matched to their own cluster and poorly matched to neighboring clusters, signifying a high-quality partition [\citenum{ikotun_cluster_2025}]. A higher Silhouette score is preferable.
	
	\item The Davies-Bouldin (DB) Index, which computes the average similarity between each cluster and its most similar one. It is defined as a minimization problem, where lower values indicate better clustering. A lower score implies that clusters are more compact and farther apart from each other [\citenum{ikotun_cluster_2025}].
	
	\item The Calinski-Harabasz (CH) Index, also known as the Variance Ratio Criterion, which measures the ratio of between-cluster to within-cluster dispersion. A higher CH score indicates denser and better-separated clusters, making it a maximization index [\citenum{ikotun_cluster_2025}].
	
\end{itemize}
		
Table \ref{tab:clusters} presents the quantitative results, averaged over 10 independent test runs to account for the stochastic nature of t-SNE. For the (CV) dataset, the metrics strongly corroborate our qualitative findings: the pseudo-synthetic autoencoder achieves the best scores across all three indices, confirming the well-defined clusters observed visually in Fig. \ref{fig:cv}(c). Conversely, the sim-only autoencoder yields the worst scores, including a negative Silhouette Index, reflecting the disorganized and overlapping latent space shown in Fig. \ref{fig:cv}(b).

The results for the augmented datasets reveal a critical insight. On the (T2) dataset, the pseudo-synthetic model achieves exceptional CVI scores, yet its classification accuracy drops to 60.78\%. This discrepancy suggests that while the model learns to form geometrically pure clusters based on low-level features introduced by the data augmentation, these clusters can become semantically confused. This underscores that optimizing for geometric cluster purity in the latent space does not guarantee improved performance on a downstream classification task, and that CVI metrics must be interpreted in conjunction with task-specific performance.
	
\begin{table*}[h!]
	\begin{minipage}{\textwidth}
		\begin{center}
			\caption{Quantitative analysis of 2D latent space cluster quality using three standard validity indices.}\label{tab:clusters}
			\begin{tabular}{@{}ccccc@{}}
				\hline
				\textbf{Real Test Dataset} & \textbf{Experiment} & \textbf{Silhouette} & \textbf{Davies-Bouldin} & \textbf{Calinski-Harabasz}\\
				\hline
				\hline
				(CV) & (1) & -0.01 & 9.48 & 0.98\\
				(CV) & (2) & -0.04 & 10.41 & 0.52\\
				(CV) & (3) & \textbf{0.10} & \textbf{6.34} & \textbf{2.51}\\
				\hline
				(T1) & (1) & 0.51 & \textbf{0.72} & 11578.54\\
			 	(T1) & (2) & \textbf{0.53} & 1.36 & \textbf{39260.67}\\
				(T1) & (3) & 0.28 & 1.64 & 3895.42\\
				\hline
				(T2) & (1) & 0.50 & 0.86 & 14538.12\\
				(T2) & (2) & 0.46 & 1.32 & 20400.30\\
				(T2) & (3) & \textbf{0.76} & \textbf{0.33} & \textbf{159172.46}\\
				\hline
			\end{tabular}
		\end{center}
	\end{minipage}
\end{table*}

\subsection{Computational Efficiency Analysis for Low-Cost Robotic Deployment}\label{subsec:efficiency}

The practical application of deep learning models in robotics is often constrained not only by predictive accuracy but also by computational efficiency, a factor of paramount importance for deployment on low-cost, resource-constrained platforms [\citenum{falaschetti_embedded_2024}]. While the preceding sections have established the high classification accuracy of our proposed policy, this analysis aims to quantitatively validate its practical viability for real-time reactive navigation. A policy, no matter how effective in theory, is of limited use if its computational and memory requirements exceed the capabilities of affordable hardware. Therefore, this analysis serves as a critical bridge between theoretical performance and real-world deployability.

To provide a realistic assessment, we eschew simple theoretical peaks and instead employ the more nuanced Roofline model as originally proposed by Williams \textit{et al.} [\citenum{williams_roofline_2009}]. The Roofline model is an insightful visual framework that bounds the attainable performance of a computational kernel by the hardware's peak throughput and its peak memory bandwidth [\citenum{williams_roofline_2009}]. We adopt its more complex formulation, which incorporates intermediate ceilings to represent the performance limitations imposed by the absence of specific software optimizations or by systemic overheads [\citenum{williams_roofline_2009}].

This approach yields a more conservative and pragmatic upper bound on performance, acknowledging that theoretical hardware limits are rarely achieved in practice. It must be emphasized that this analysis provides an estimation of performance. While grounded in a rigorous model, the precise values of these ceilings can only be determined through exhaustive empirical micro-benchmarking on the physical hardware.

The evaluation is conducted across a curated set of hardware platforms representative of the low-cost robotics domain. These include a standard Linux laptop, which serves as a baseline for development, the Raspberry Pi 3 Model B+\footnote{https://datasheets.raspberrypi.com/rpi3/raspberry-pi-3-b-plus-product-brief.pdf}, selected for its ubiquity, and the Raspberry Pi 4 Model B\footnote{https://datasheets.raspberrypi.com/rpi4/raspberry-pi-4-datasheet.pdf} and NVIDIA Jetson Nano 2 GB\footnote{https://cdn.sparkfun.com/assets/7/9/7/b/d/Jetson\_Nano\_2GB\_datasheet.pdf}, chosen for their widespread adoption. The key technical specifications for these platforms, compiled from official datasheets and verified sources, are summarized in Table \ref{tab:platforms}.

\begin{table*}[h!]
	\begin{minipage}{\textwidth}
		\begin{center}
			\caption{Technical specifications of evaluated hardware platforms.}\label{tab:platforms}
			\begin{tabular}{@{}cccc@{}}
				\hline
				\textbf{Platform} & \textbf{Processor} & \textbf{Memory} & \textbf{AI Accelerator}\\
				\hline
				\hline
				Linux Laptop & \makecell{Intel Core i7-8750H \\ 6-core @ 2.20 GHz} & 32 GB DDR4 & \makecell{NVIDIA GeForce \\ RTX 2080}\\
				\hline
				Raspberry Pi 3B+ & \makecell{ARM Cortex-A53 \\ Quad-core @ 1.40 GHz} & 1 GB LPDDR2 & N/A\\
				Raspberry Pi 4B & \makecell{ARM Cortex-A72 \\ Quad-core @ 1.50 GHz} & 4 GB LPDDR4 & N/A\\
				Jetson Nano 2 GB & \makecell{ARM Cortex-A57 \\ Quad-core @ 1.43 GHz} & 2 GB LPDDR4 & \makecell{128-core NVIDIA \\ Maxwell}\\
				\hline
			\end{tabular}
		\end{center}
	\end{minipage}
\end{table*}

\subsubsection{Theoretical Complexity of the Proposed Architecture}\label{subsubsec:complexity}

A bottom-up, hardware-agnostic analysis of the proposed architecture establishes a baseline of its computational and memory requirements. The complete pseudo-synthetic architecture consists of a convolutional encoder, a shared convolutional decoder, and a lightweight MLP classifier. The parameter counts for each component, as measured using the TensorFlow Keras framework, are as follows:

\begin{itemize}
	
	\item \textbf{VAE Encoder}: 2172864 parameters.
	
	\item \textbf{Shared VAE Decoder}: 1302915 parameters.
	
	\item \textbf{Full VAE Model}: 3475779 parameters.
	
	\item \textbf{MLP Classifier}: 3704 parameters.
	
\end{itemize}

For the inference phase, the decoder is not utilized. The deployed model consists only of the convolutional encoder and the MLP classifier. This results in an operational inference model with a total of 2176568 parameters (\textit{i.e.}, $\sim$2.18 million).

The key metrics for our efficiency analysis are derived from this inference model. The memory footprint is calculated based on the parameter count, assuming standard 32-bit Floating-Point (FP32) precision, where each parameter occupies 4 bytes. Therefore, the model's memory footprint corresponds to 8.31 MiB. The computational cost is measured in Floating-Point Operations (FLOPs), following the common convention that a Multiply-Accumulate (MAC) operation is equivalent to approximately 2 FLOPs. Based on the model's layer-by-layer structure, we estimate its computational load for a single forward pass on a 64x64 RGB input image to be $\sim$169.4 MFLOPs.

\subsubsection{Performance Estimation via the Roofline Model}\label{subsubsec:performance}

The Roofline model establishes an attainable performance bound based on the interplay between the model's Operational Intensity (OI) and the hardware's peak capabilities. According to Williams \textit{et al.} [\citenum{williams_roofline_2009}], the attainable performance is given by the equation:

\begin{equation}
	\begin{split}
		Attainable \enspace GFLOPs/s = \min(&Peak \enspace Floating \text{-} Point\\ & Performance,\\
		&Peak \enspace Memory \enspace Bandwidth\\ & \times Operational \enspace Intensity).\label{eqn:performance}
	\end{split}
\end{equation}

From this, we derive the estimated latency and the throughput, measured in Frames Per Second (FPS). Our model exhibits an operational intensity of $\sim$19.33 FLOPs/Byte, indicating a high ratio of computation to memory access.

To provide a more pragmatic estimation, we apply the complex Roofline model, incorporating intermediate ceilings that represent performance under sub-optimal conditions (\textit{e.g.}, with system overheads). Table \ref{tab:performance} presents these more conservative estimations.

\begin{table*}[h!]
	\begin{minipage}{\textwidth}
		\begin{center}
			\caption{Roofline-based performance estimation for the proposed pseudo-synthetic model.}\label{tab:performance}
			\begin{tabular}{@{}ccccc@{}}
				\hline
				\textbf{Platform} & \makecell{\textbf{Memory} \\ \textbf{(MiB)}} &  \makecell{\textbf{Cost} \\ \textbf{(MFLOPs)}} & \makecell{\textbf{Latency} \\ \textbf{(ms)}} & \makecell{\textbf{Throughput} \\ \textbf{(FPS)}}\\
				\hline
				\hline
				Linux Laptop (CPU) & 8.31 & 169.40 & $\sim$1.10 & $\sim$900\\
				\hline 
				Raspberry Pi 3B+ (CPU) & 8.31 & 169.40 & $\sim$257 & $\sim$4\\
				Raspberry Pi 4B (CPU) & 8.31 & 169.40 & $\sim$21.70 & $\sim$46\\
				Jetson Nano 2 GB (GPU) & 8.31 & 169.40 & \textbf{$\sim$1.20} & \textbf{$\sim$833}\\
				\hline
			\end{tabular}
		\end{center}
	\end{minipage}
\end{table*}

The analysis presented in Table \ref{tab:performance} confirms the suitability of our proposed policy for deployment in low-cost robotics for real-time, reactive visual navigation. The model's low memory footprint and modest computational load are well within the capabilities of modern embedded systems.

The Linux laptop serves as a strong baseline, indicating that development and testing can be performed with extremely high throughput. The Raspberry Pi 3B+, however, is shown to be a challenging platform. Our realistic estimate of $\sim$4 FPS falls below the typical 30-60 FPS range considered sufficient for smooth navigation, suggesting this hardware is unsuitable for the task without a dedicated accelerator.

This estimation aligns with the findings of Feng \textit{et al.} [\citenum{feng_improving_2021}], who measured an inference time of 117.7 ms (\textit{i.e.}, $\sim$8.5 FPS) for their VAE-based model, which relies on 2.2 million parameters, on the same hardware. Given our inference model has a very similar parameter count, both results confirm that this class of deep learning model is too demanding for the Raspberry Pi 3B+ alone.

In contrast, the Raspberry Pi 4B emerges as a highly viable platform. Its estimated performance of $\sim$46 FPS comfortably falls within the 30-60 FPS range required for real-time applications, enabling smooth and reactive control. This result confirms that the generational leap in performance from the Raspberry Pi 3B+ is sufficient to make it a primary candidate for deploying our navigation policy.

The NVIDIA Jetson Nano 2 GB provides exceptional performance, with an estimated throughput of $\sim$833 FPS. Even when accounting for significant system overheads that would reduce this number in practice, the performance is far beyond the requirements of reactive navigation. This computational margin makes it suitable for running more complex tasks in parallel, such as mapping or object recognition.

To further contextualize these results, we compare them with the work of Falaschetti \textit{et al.} [\citenum{falaschetti_embedded_2024}], who evaluated a compressed Tiny You-Only-Look-Once (YOLO) v3 variant for vehicle and pedestrian detection. Their final model has a slightly lower parameter count of $\sim$1.92 million but a significantly higher computational load: 948 MFLOPs. Furthermore, it operates on larger 416x416 images.

On the Jetson Nano 2 GB, Falaschetti \textit{et al.}'s model achieved an empirical 36.46 FPS. Our model is estimated to be substantially faster primarily due to two intentional design trade-offs: a much lower computational load and a significantly smaller input image resolution of 64x64. This highlights a key principle for vision-based robotics: by focusing the network on learning from compressed, low-dimensional representations tailored to the navigation task, the computational burden of the policy can be dramatically reduced compared to more general-purpose object detection architectures. This makes our approach particularly well-suited for the target application domain of real-time reactive navigation on resource-constrained hardware.

\subsection{Case Study: Real-World Deployment for Reactive Collision-Free Exploration}\label{subsec:deployment}
	
To empirically validate the practical applicability and real-time performance of our proposed transfer learning method, we deployed the learned image classification policy on a physical robotic platform. This subsection details the experimental configuration and demonstrates that our pseudo-synthetic policy, developed through a hybrid strategy of domain randomization and domain adaptation, effectively translates into a robust, reactive navigation behavior capable of performing indoor exploration tasks.

\subsubsection{Experimental Setup}\label{subsubsec:setup}

The primary goal of this case study was to validate the navigational competence of the proposed policy, separating its effectiveness from hardware performance limitations. The only embedded platform available for deployment was the Raspberry Pi 3 Model B+, for which our efficiency analysis in Subsection \ref{subsec:efficiency} estimated a throughput of merely $\sim$4 FPS. This performance level is insufficient for the demands of real-time reactive navigation. Given that more powerful embedded platforms, such as the Raspberry Pi 4 Model B or NVIDIA Jetson Nano 2 GB, and compatible hardware accelerators were not available, a direct deployment on the Raspberry Pi 3B+ was deemed infeasible.

Consequently, we adopted a distributed architecture that leverages our Linux laptop as the primary computational node for high-level inference. This decision directly links our computational analysis to our empirical validation, demonstrating a practical methodology for testing complex policies when target hardware is constrained. As predicted by the analysis in Subsection \ref{subsec:efficiency}, the laptop's processing power was more than sufficient to execute the inference model in real-time.

\begin{figure}[h!]
	\centering
	\includegraphics{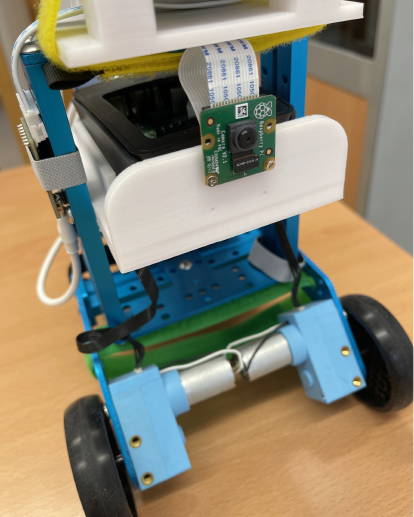}
	\caption{Low-cost physical robot used for real-world deployment. The platform is a customized Makeblock mBot Ranger equipped with a single forward-facing monocular RGB camera module, which provides all perceptual information for the navigation task.}\label{fig:robot}
\end{figure}

The final hardware configuration, shown in Fig. \ref{fig:robot}, consisted of a customized Makeblock mBot Ranger robot, a differential-drive agent [\citenum{stefek_energy_2020}]. For perception, the robot was equipped with a forward-facing Raspberry Pi Camera Module 2\footnote{https://www.raspberrypi.com/products/camera-module-v2/}. For low-level actuation, an on-board Raspberry Pi 3B+ was connected via USB serial port to an Arduino Leonardo micro-controller\footnote{https://docs.arduino.cc/hardware/leonardo/}. The control pipeline operated as follows:

\begin{enumerate}
	
	\item The robot's camera streamed RGB images to the Linux laptop via the Robot Operating System (ROS) middleware.
	
	\item On the laptop's CPU, the VAE encoder and MLP classifier processed the images to predict a navigation command (\textit{i.e.}, move forward, turn right, turn left, or stop).
	
	\item This command was translated into a specific linear and rotational velocity pair, $V = \{(v_{linear}, v_{rotational})\}$, and sent back to the robot's Raspberry Pi 3B+ via ROS. Based on the method from [\citenum{stefek_energy_2020}], these velocities were used to calculate the precise speed for the agent's left and right motors. This calculation, which requires the distance $\mathbb{D}$ between the wheels, is performed as:
	\begin{equation}
	v_{left} = v_{linear} + \mathbb{D} \frac{v_{rotational}}{2}.\label{eqn:left}
	\end{equation}
	\begin{equation}
	v_{right} = v_{linear} - \mathbb{D} \frac{v_{rotational}}{2}.\label{eqn:right}
	\end{equation}
	
	\item The Raspberry Pi 3B+ relayed the commands to the Arduino Leonardo micro-controller, which was responsible for the low-level control of the robot's two independent motors.
	
\end{enumerate}

This configuration successfully isolated the policy's decision-making process from the hardware limitations of the robot, allowing the laptop to handle the computational load effortlessly. As a result, the robot executed its reactive navigation task in the real-world corridor without any performance-related issues, enabling a pure validation of the policy's effectiveness.

\subsubsection{Navigational Trajectory Analysis}\label{subsubsec:trajectory}

The core of this validation involved transferring the pseudo-synthetic deep neural network architecture to the physical robot to perform an autonomous, map-less exploration of the real indoor corridor. Deployed from a central starting position, our robot successfully navigated a significant portion of the environment by reactively avoiding a variety of obstacles.

\begin{figure*}[h!]
	\centering
	\includegraphics[width=\textwidth]{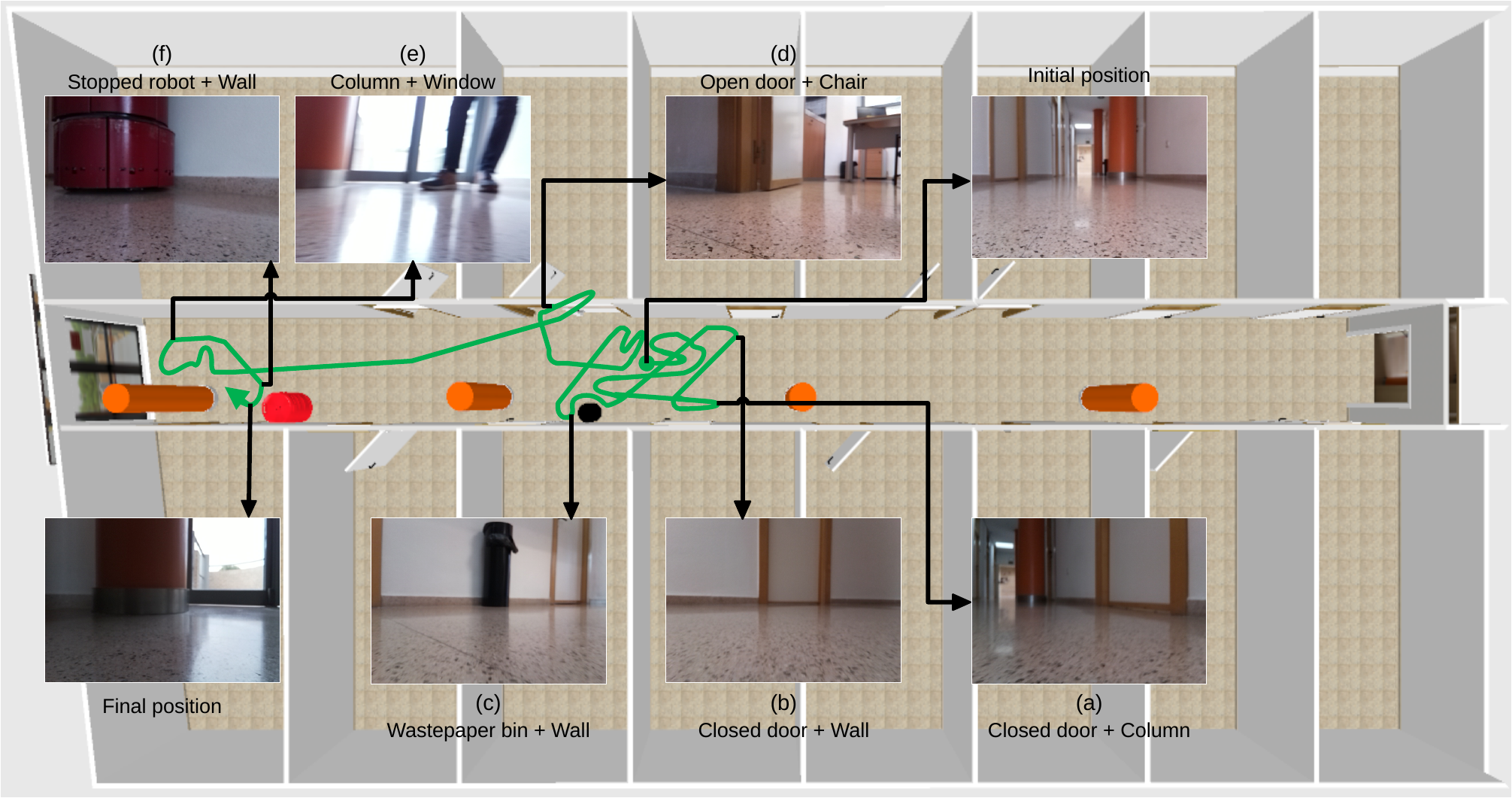}
	\caption{Real-time evaluation of the proposed image classification policy on a low-cost physical robot. The green line depicts the path taken by the differential-drive agent during an autonomous, map-less exploration task in our indoor corridor, located in the Department of Computer Science and Artificial Intelligence at the University of Alicante. The trajectory demonstrates the robot's ability to robustly navigate and avoid distinct obstacles, analyzed at six different checkpoints. (a) A closed door followed by a column. (b) A closed door placed in a side wall. (c) A wastepaper bin located in front of a wall. (d) An open door and a chair corresponding to an office. (e) A column placed beside a window. (f) A robot stopped on a wall.}\label{fig:trajectory}
\end{figure*}

Fig. \ref{fig:trajectory} illustrates the trajectory followed by the mobile agent, which was reconstructed from the sequence of executed motor commands. The path highlights the reactive nature of our navigation system. Its trajectory is not pre-planned, but rather an emergent behavior dictated by the continuous perception of the environment. We identified six critical checkpoints along this path to analyze the robot's real-time decision-making process in challenging scenarios.
	
\subsubsection{Reactive Behavior at Critical Checkpoints}\label{subsubsec:obstacles}
	
The robot's behavior at the checkpoints detailed in Fig. \ref{fig:obstacles} reveals a high level of navigational competence, directly attributable to the policy learned via our hybrid transfer learning approach. The sequence of commands executed in response to the perceived images is described as follows:

\begin{figure*}[h!]
	\centering
	\includegraphics[height=0.82\textheight, width=\textwidth]{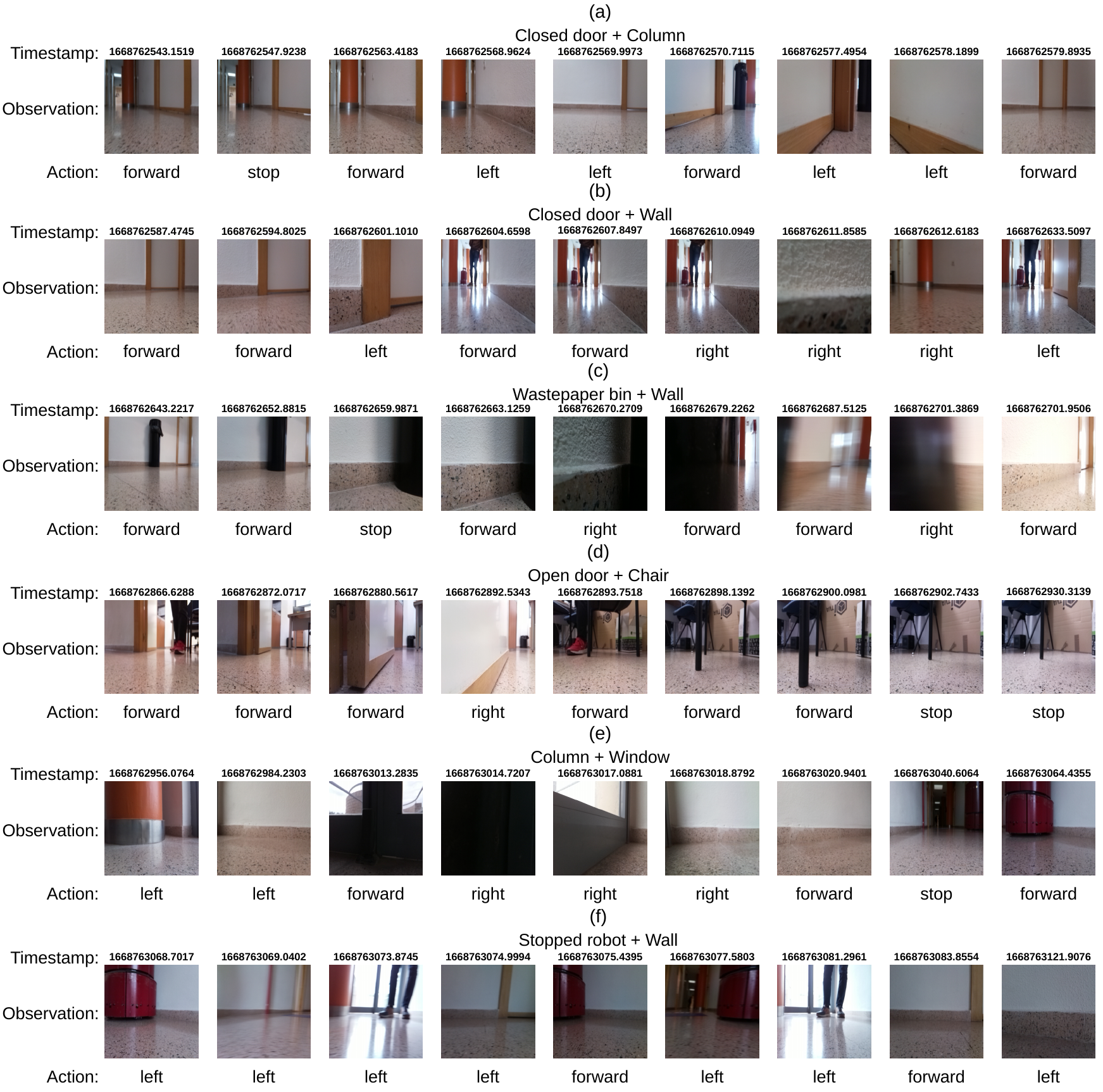}
	\caption{Detailed analysis of the robot's reactive behavior at six different checkpoints. These examples illustrate successful collision avoidance with challenging combinations of obstacles with diverse shapes and textures under varying illumination conditions. (a) A closed door followed by a column. (b) A closed door placed in a side wall. (c) A wastepaper bin located in front of a wall. (d) An open door and a chair corresponding to an office. (e) A column placed beside a window. (f) A robot stopped on a wall.}\label{fig:obstacles}
\end{figure*}

\begin{enumerate}[(a)]
	
	\item \textbf{A closed door followed by a column.} The agent initially moved alongside the door. Upon approaching the column, it executed a decisive half-turn, demonstrating effective obstacle avoidance. It subsequently proceeded to the opposite side of the corridor, ensuring exhaustive area coverage without collision.
	
	\item \textbf{A closed door placed in a side wall.} The robot headed towards the door and made a timely left turn to avoid its frame. It then followed the wall before maneuvering towards the central area of the corridor, having fully visited the sector.
	
	\item \textbf{A wastepaper bin located in front of a wall.} The robot cautiously approached the wall, turned right to avoid it, and then successfully maneuvered around the wastepaper bin. After passing the obstacle, it turned again to move parallel to the side wall.
	
	\item \textbf{An open door and a chair corresponding to an office.} Through a succession of forward movements and a decisive right turn, the robot entered the office. It advanced until it reached a chair and positioned itself underneath, pausing for several time steps to cautiously handle this complex scenario.
	
	\item \textbf{A column placed beside a window.} The agent performed several turns to explore a windowed area without collision. It then traveled parallel to the adjacent wall and successfully turned around upon reaching one end of the corridor, proving its ability to handle multiple obstacles of diverse nature.
	
	\item \textbf{A robot stopped on a wall.} The agent perceived a large, stationary robot at a distance and executed a sequence of left turns to perform a complete rotation. This maneuver allowed it to explore a new sector while avoiding collision with both the other robot and the nearby wall.
	
\end{enumerate}

\subsubsection{Robustness to Environmental Variations}\label{subsubsec:variations}

\begin{figure*}[h!]
	\centering
	\includegraphics[width=\textwidth]{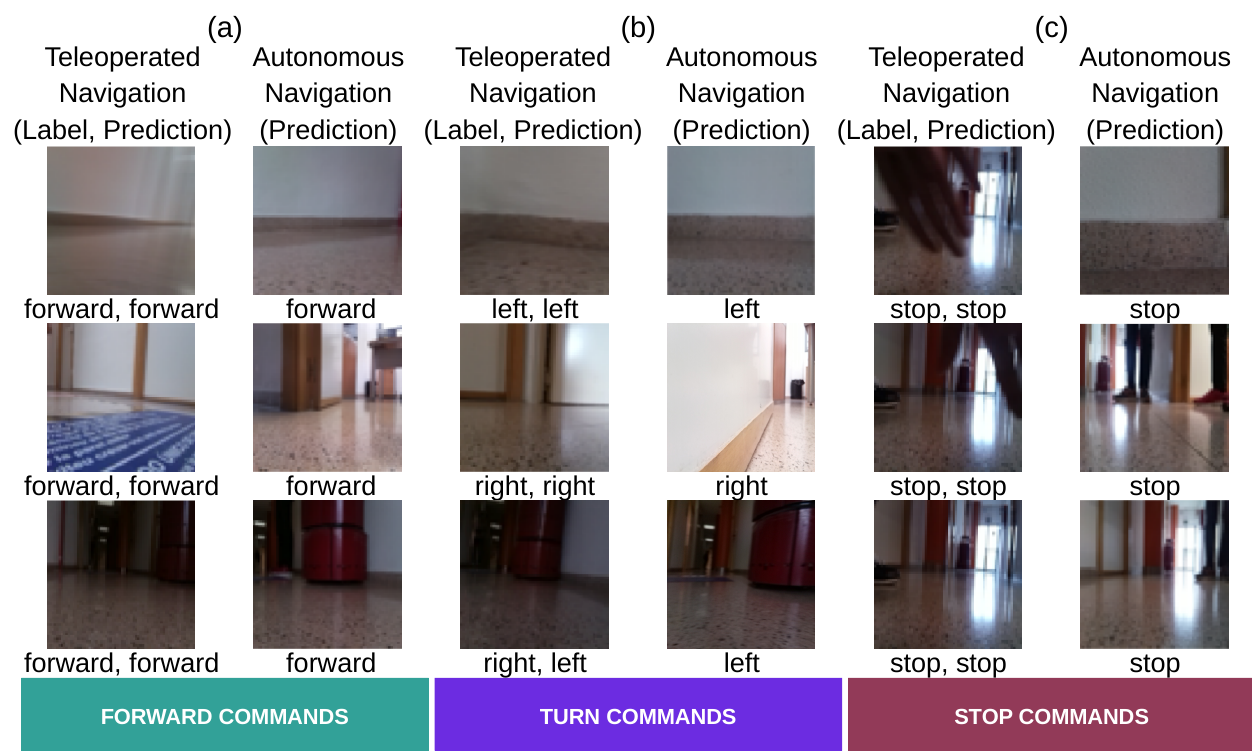}
	\caption{Qualitative evaluation of the proposed image classification policy's generalization, adaptation, and robustness. We compared teleoperated navigation with fully autonomous navigation on two different days with varying conditions. Our reactive system demonstrates consistent behavior in response to several unknown situations, successfully adapting to lighting changes, novel objects, and dynamic obstacles. (a) Forward scenarios. (b) Turn scenarios. (c) Stop scenarios.}\label{fig:variations}
\end{figure*}

To further assess the generalization and adaptation capabilities of the proposed policy, we conducted a qualitative comparison of the system's performance on two different days, under varying lighting conditions and with the presence of novel elements not seen during training. As shown in Fig. \ref{fig:variations}, our autonomous navigation policy, depicted in the right column of each panel, consistently produced coherent and safe actions that mirrored an expert-demonstrated (\textit{i.e.}, teleoperated by a human expert) behavior. The latter is represented in the left column of each panel.

As illustrated in Fig. \ref{fig:variations}(a), our navigation system demonstrated robustness to environmental variations, such as changes in ambient light, the presence or absence of certain objects (\textit{e.g.}, the blue adhesive sign on the floor), and the appearance of a person, correctly predicting forward commands in open spaces. These commands encourage greater exploration of the environment.

Most notably, the proposed policy exhibited intelligent reactivity. For example, in the last scenario of Fig. \ref{fig:variations}(b), which is analogous to a case in Fig. \ref{fig:predictions}(b), the ground-truth label from the expert corresponded to a right turn, but our policy predicted a left turn. This decision was still correct in a reactive context, as the crucial action was to turn and avoid the imminent obstacle, not to replicate the exact direction. This behavior, inspired by biological stress reactions, prioritizes immediate safety over exact imitation, a key feature for robust embodied navigation in the real world.

Finally, as depicted in Fig. \ref{fig:variations}(c), the policy proved adaptable to dynamic events, correctly issuing stop commands when a person's hand or feet suddenly entered the camera's field of view, thus preventing collisions in dynamic situations. This comprehensive real-world validation confirms that our hybrid transfer learning approach produces policies that are not only accurate but also effective, robust, and flexible when deployed on low-cost physical robots in challenging indoor scenarios.

\section{Conclusions}\label{sec:conclusions}

This study has presented and validated a hybrid transfer learning method designed for vision-based robotic navigation, which enables the efficient training of classification policies using primarily simulated data supplemented with limited real-world examples. The core of our approach is a dual convolutional variational autoencoder architecture with a shared decoder, which integrates domain randomization with a subsequent, refined domain adaptation stage.

The contributions of this work fulfill a dual objective: first, we validated the practical efficacy of the proposed policy through a proof-of-concept deployment on a low-cost physical robot performing a reactive, collision-free exploration task. Second, we provided a rigorous estimation of the model's computational performance, confirming its suitability for deployment on resource-constrained embedded platforms.

Our comprehensive experimental results substantiate our three foundational hypotheses:

\begin{enumerate}[1.]
	
	\item Training a policy exclusively with a small real-world dataset is insufficient for achieving robust performance in autonomous navigation tasks. This real-only approach yielded an average success rate of approximately 70\% on unseen real-world data, a limitation that stems from the inherent difficulties in acquiring extensive and balanced labeled datasets from real-world environments.
	
	\item A policy trained solely on an extensive synthetic dataset via domain randomization exhibits limited real-world generalization, failing to surpass a 41\% average success rate on real test images. This finding underscores that systematic discrepancies between simulation and reality persist even when using a high-fidelity simulator, making direct sim-to-real transfer unreliable.
	
	\item Our proposed hybrid transfer learning method significantly improves predictive capacity, achieving an average success rate of nearly 91\% on unseen real-world data. This is accomplished by first leveraging domain randomization and subsequently performing feature-level domain adaptation, a process initiated with a small set of real images and enhanced via two complementary data augmentation techniques. The dual VAE architecture, with its shared decoder, effectively forces the alignment of feature distributions, thereby bridging the simulation-to-reality domain gap.
	
\end{enumerate}

Building on these findings, our future research will proceed along three primary directions. Foremost, we will conduct an empirical validation of the computational performance estimations presented in Subsection \ref{subsec:efficiency}. This will involve deploying and benchmarking the inference model on the target low-cost hardware, including the Raspberry Pi 4 Model B and the NVIDIA Jetson Nano 2 GB, to obtain precise latency and throughput metrics.

Second, we intend to expand the real-world evaluation phase by subjecting the navigation policy to more complex and dynamic indoor environments. These experiments will assess the policy's robustness against greater environmental variations, including significant changes in illumination, higher levels of clutter, and the presence of dynamic obstacles.

Finally, we plan to extend the framework to swarm robotics [\citenum{brambilla_swarm_2013}] platforms. This involves deploying our VAE-based perception model across multiple low-cost agents, enabling decentralized, real-time visual processing for collaborative navigation tasks such as coordinated exploration, target tracking, and surveillance in large-scale, complex real-world scenarios. 

\section*{CRediT author statement}

\textbf{\'{A}lvaro D\'{i}ez:} Conceptualization, Methodology, Software, Validation, Formal analysis, Investigation, Data curation, Writing - Original draft, Writing - Review and editing, Visualization. \textbf{Fidel Aznar:} Conceptualization, Methodology, Data curation, Validation, Resources, Supervision, Project administration.

\section*{Declaration of interests}

The authors declare that they have no known competing financial interests or personal relationships that could have appeared to influence the work reported in this paper.

\section*{Declaration of generative AI and AI-assisted technologies in the writing process}

During the preparation of this work, the authors used Google Gemini 2.5 in order to improve the quality and readability of the written text. After using this tool, the authors reviewed and edited the content as needed and take full responsibility for the content of the published article.

\section*{Acknowledgments}

This work was supported by the Spanish Ministry of Science, Innovation and Universities under Grant PID2025-169299OB-I00.

\end{document}